\documentclass[review]{elsarticle}

\usepackage{siunitx,booktabs,multicol,multirow,geometry,pdflscape,amsfonts,amsmath}
\usepackage{hyperref}
\usepackage{subcaption}
\usepackage{graphicx}
\usepackage{xcolor}
\usepackage{enumitem}
\usepackage[T1]{fontenc}
\usepackage{longtable}
\usepackage{algorithm}
\usepackage{algpseudocode}

\journal{Journal of \LaTeX\ Templates}

\begin{document}

\begin{frontmatter}

\title{Applying Autonomous Hybrid Agent-based Computing\\ to Difficult Optimization Problems}

\author{Mateusz Godzik, Jacek Dajda, Marek Kisiel-Dorohinicki, Aleksander Byrski}
\address{AGH University of Science and Technology, Al. Mickiewicza 30, 30-059 Krakow, Poland, \{godzik,dajda,doroh,olekb\}@agh.edu.pl}

\author{Leszek Rutkowski}
\address{Institute of Computer Science, AGH University of Science and Technology}

\author{Patryk Orzechowski, Joost Wagenaar}
\address{Department of Biostatistics, Epidemiology and Informatics, University of Pennsylvania}

\author{Jason H. Moore}
\address{Department of Computational Biomedicine at Cedars-Sinai Medical Center}

% %% or include affiliations in footnotes:
% \author[mymainaddress,mysecondaryaddress]{Elsevier Inc}

% \author[mysecondaryaddress]{Global Customer Service\corref{mycorrespondingauthor}}
% \cortext[mycorrespondingauthor]{Corresponding author}

% \address[mymainaddress]{1600 John F Kennedy Boulevard, Philadelphia}
% \address[mysecondaryaddress]{360 Park Avenue South, New York}

\begin{abstract}
Evolutionary multi-agent systems (EMASs) are very good at dealing with difficult, multi-dimensional problems, their efficacy was proven theoretically based on analysis of the relevant Markov-Chain based model. Now the research continues on introducing autonomous hybridization into EMAS. This paper focuses on a proposed hybrid version of the EMAS, and covers 
selection and introduction of a number of hybrid operators and defining rules for starting the hybrid steps of the main algorithm. Those hybrid steps leverage existing, well-known and proven to be efficient metaheuristics, and integrate their results into the main algorithm.
The discussed modifications are evaluated based on a number of difficult continuous-optimization benchmarks.
\end{abstract}

\begin{keyword}
Agent-based computing \sep Hybrid metaheuristics \sep Nature-inspired algorithms
\end{keyword}

\end{frontmatter}

%\linenumbers

\section{Introduction}
Despite the recent increases in computer performance, not all problems can be resolved in a timely manner. Solving problems by deterministic algorithms often takes too long (e.g., exceeding a dozen or so of cities in the traveling salesman problem (TSP) is such an example \cite{tspcomplexity}). However, some problems do not have deterministic solutions. In these as well as in other cases, novel stochastic methods can help. If other solutions do not meet the assumed goals, one can use metaheuristics; their great advantage is that they do not require information about the characteristics of the search space. One of the advantages is the ability to tune the algorithm through the selection of parameters (cf. iRace \cite{LOPEZIBANEZ201643}) or the ability to combine it with other algorithms to create hybrid algorithms (cf. Talbi \cite{talbitaxonomy}). We are still looking for new metaheuristics, as it is impossible to find a single method that will solve all problems with the same accuracy (cf. Wolpert and MacReady \cite{wolpert_pages}).

%An example of such an algorithm is \emph{Evolutionary Multi-Agent System}),(EMAS), which has been with us since 1996 \cite{kc_pages}.
An example of such metaheuristics is the concept of an evolutionary multi-agent system (EMAS), which was introduced in 1996 \cite{kc_pages}.
It is a kind of combination of the evolutionary method \cite{goldberg} with the agent paradigm \cite{softwareagents}, resulting in a program in which agents are part of the computational process that searches the admissible space. Moreover, it leverages e.g., a decentralized selection mechanism based on a non-renewable resource and two actions: reproduction, and death. There is no centralized control, so such algorithms can be easily implemented in a concurrent manner; this allows for a significant reduction in the computation time.
As was proven in \cite{biul1} (a formal proof that followed the works of Michael Vose \cite{vose} showed that a Markov-chain that models the dynamics of EMAS is ergodic, so this algorithm can reach any possible solution of a given problem), it has become a solid base for attempts to be combined with other algorithms; this has yielded interesting hybrid computing methods (e.g., \cite{GodzikGPSBK19}). 
%(see, e.g. \cite{PlaczkiewiczSSP18,GodzikGPSBK19}).
%However, hybridizing EMAS with other algorithms, one may encounter the problem of redistributing the energy of agents making their own decisions to use algorithms that do not use energy (\cite{ICCCI2020}). This article looks at several solutions to this problem and carefully examines the best of them.

This grounds a hybrid evolutionary multi-agent systems (HEMAS) and discusses the selected classic metaheuristics in detail, which have become the bases for the hybridization of EMAS. These metaheuristics are executed during the main course of the EMAS algorithm (when certain conditions are met). Moreover, these conditions are also discussed, and all the deliberations are illustrated with the outcomes of experiments that are based on difficult continuous optimization problems. 

The main contributions, novelties, and characteristics of this paper can be summarized as follows:
\begin{itemize}
    \item We propose a novel concept of an agent-based computing system for solving difficult optimization problems.
\item Our proposition involves hybridizing classical EMAS systems with global optimization metaheuristic algorithms -- particle swarm optimization (PSO), and the genetic algorithm (GA) -- leading to HEMAS systems.
\item The hybridization is triggered by implementing several rules that have been proposed to design the HEMAS systems.
\item  We solved the problem of redistributing the agents' energy by using an appropriate redistribution operator.
\item  Through extensive computer simulations, we show that the proposed HEMAS computing system produces competitive results as compared to EMAS (the version with two hybrid algorithms was better than the one with one hybrid algorithm, yet the version with three hybrid algorithms was not superior to the one with two) by using proper comparison metrics (the numbers of the evaluations of the fitness functions); these are independent from the algorithm, implementation, and hardware.
\end{itemize}

The outline of the paper is as follows. After the introduction, we describe the original EMAS (with references to the state of the art) and follow with the description of HEMAS. Then, the details that are related to the hybridization with the selected metaheuristics are given (the concept, and the later parameters). Finally, the experimental setting and the results are shown and discussed, and the paper is concluded.

\section{Original EMAS}
\begin{figure}[ht]
\centering
\includegraphics[scale=.25]{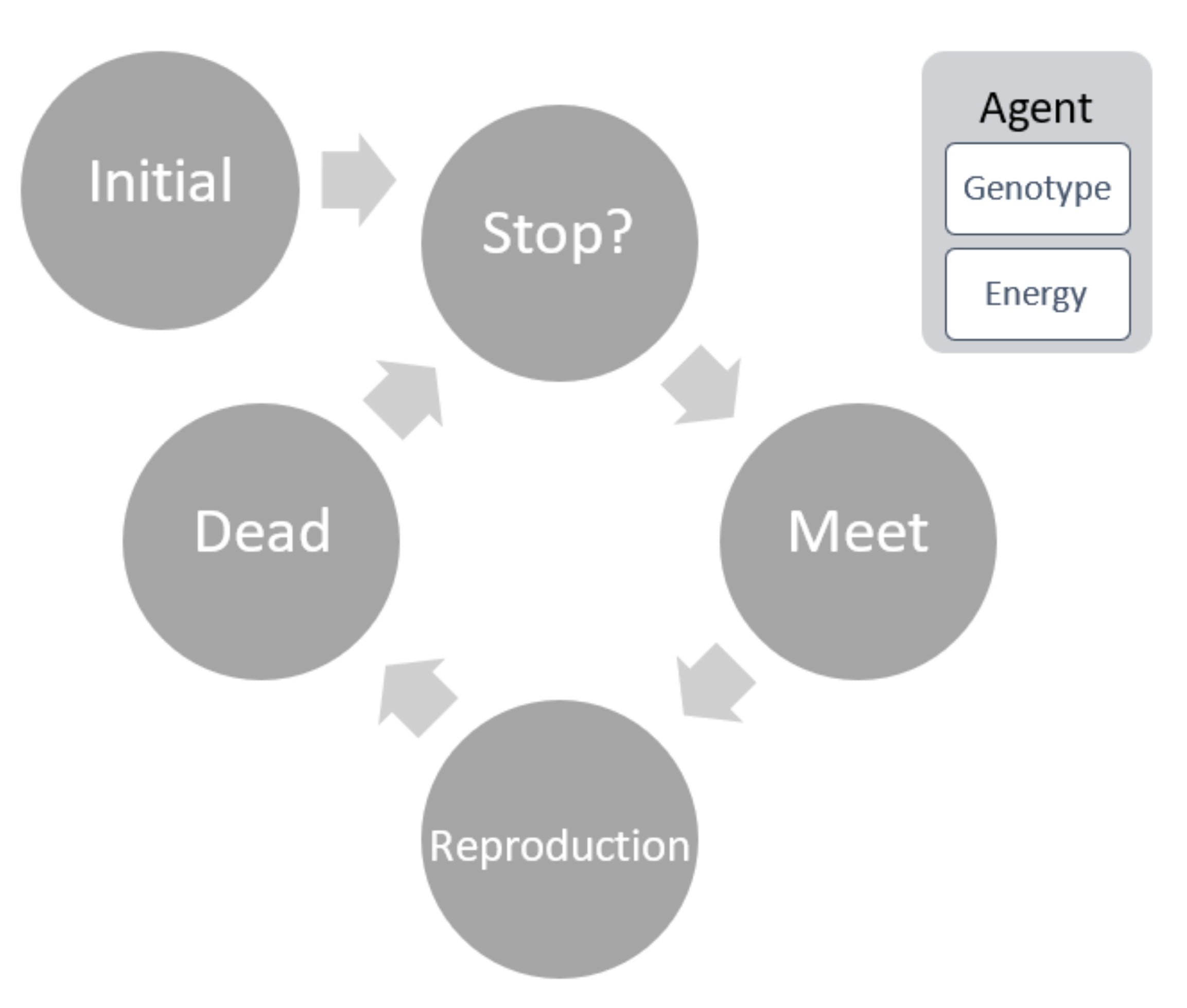}
\caption{Diagram of evolutionary multi-agent systems}
\label{rys:emas}
\end{figure}

EMAS may be perceived as a ``proactive'' alternative to classical
evolutionary computation techniques %\cite{compint}
that can relieve evolutionary metaheuristics of several inconsistencies with a real-life evolution; e.g., a lack of global control, and asynchronous reproduction. In this system, solutions (genotypes) are assigned to agents, realizing the several types of actions that are available to them in order to upgrade their solutions. 
Agents can meet with each other and either compete or exchange resources. In the former case, only the richer agent is allowed to reproduce (analogical to selection in EA); in the latter, part of the resources of the poorer agent are allocated to the richer one (analogical to crossover in EA) \cite{kc_pages}. For a schematic view of EMAS, one can refer to Fig. \ref{rys:emas}. It is to note that the correctness of EMAS as a global universal optimizer has been formally proven by using Markov chain-based models that were inspired by the theoretical works of Michael Vose \cite{form1}. %\cite{form1,form2}.
EMAS also has many extensions; e.g., an immunological one \cite{immuno} that was applied to solve different single-criterion and multi-criteria problems.

The relationship between the elements of the algorithm is presented in Fig. \ref{rys:emas}. The agency is based on the particular implementation that has been used in numerous projects over the last two decades, and the agents are implemented as entities that undergo event-driven simulations \cite{des}; ergo, the focus can be placed on the actual algorithm and not waste time on implementing the agents as threads or processes, along with all of the necessary communication mechanisms.

The detailed EMAS algorithm is summarized in Algorithm \ref{alg:emas}. It can be noticed that the setting is quite similar to the classic genetic algorithm, for example. This implementation method of an agent-based computing system makes it possible to retain the idea of agency yet utilize simple technologies (in the discussed case, the system is based on jMetal \cite{jmetal}). The agent-oriented activities are actually realized inside the steps; e.g., \textit{meetStep()}, \textit{reproStep()}, and \textit{deadStep()}. The event-driven simulation approach consists of making it possible that each agent realizes these steps one at a time when certain conditions are true. Thus, the concurrency is simulated, and the whole approach is very easy to implement and port among different metaheuristic-oriented software frameworks.

The event-driven approach allows for the building of a system that consists of loosely coupled components that communicate by using a simple mechanism of transmitted and received events. This architecture considerably simplifies the system distribution and makes it more scalable and resistant to failures. One can notice that these are expected features for each computing architecture. It is also compatible with the agent-computing paradigm that assumes that loosely coupled independent agents communicate with each other. The implementation of the agent paradigm in the event-driven architecture can be easily realized by treating agents as both emitters and consumers of the messages and can be freely configured and adopted to different hybrids of EMAS.

\begin{algorithm}[ht]
\caption{EMAS algorithm pseudocode}\label{alg:emas}
\begin{algorithmic}[1]
    \Function{run}{}
        \State initProgress()
        \State createInitialPopulation()

        \While{\textbf{!}isStoppingConditionReached()}
            \State meetStep()
            \State reproStep()
            \State deadStep()
            \State updateProgress()
        \EndWhile
    \EndFunction
\end{algorithmic}
\end{algorithm}

\section{Agent-based hybrid EMAS}
EMAS also has many extensions that can be used to solve different single-criterion and multi-criteria problems. Among others, it is worth mentioning immunological EMAS, elitist EMAS for multi-objective optimization, CoEMAS, memetic agent-based continuous optimization, and hybrids of EDA-type and EMAS algorithms. All of these are briefly presented below.

Immunological EMAS (iEMAS) \cite{immuno} assumes the introduction of a new group of agents that act as lymphocyte T-cells (following an immunological inspiration). The goal of these special agents is to introduce a negative-selection mechanism into the evolution process that would eliminate other agents with similar genotypes. This can be achieved by either killing or weakening a selected agent by decreasing its strength. For this purpose, a defined matching function is used to calculate the similarity of a tested agent to a lymphocyte-agent (e.g., the percentage of similar genes). Lymphocyte agents are created after the death of an agent and are given a mutated variant of its genotype. Over a period of time, the lymphocyte patterns that recognize ``good'' agents (possessing high amounts of energy) are eliminated. In this way, the evolution promotes those lymphocyte agents that bear a resemblance to ``bad'' agents so that they can be continuously removed from the system -- thus leaving the ``good'' agents intact. This approach is useful for applications in which the fitness evaluation is time-consuming, as the operation of lymphocyte agents is less expensive.

The elitist evolutionary multi-agent system for multi-objective optimization is another hybrid of EMAS that was introduced in \cite{SIWIK2008}. The main idea behind this was to create a ranking of the best solutions according to an additional feature called ``prestige.'' Each agent starts its life with a prestige level of zero; then, it increases its prestige level by meeting and comparing itself with other agents using the fitness function. This resource is accumulated by the agent throughout its lifetime and cannot be lost. The second important assumption of this approach is the introduction of a special elite island from where the agents cannot migrate to other islands (and in this way cannot take part in the ongoing evolution process). To migrate to this island, an agent needs to possess a proper level of prestige. The elitist approach allows for a deeper exploration of a solution's frontier and, therefore, is suitable for multi-objective optimization. Another advantage is its elegance and simplicity as well as its ease in providing additional extensions.

The problem with classical evolutionary algorithms and EMAS-type systems is the low diversity in their populations. This is a drawback in the context of solving multi-modal optimization problems where one needs to create species that are localized around different local optima. These problems can be solved by using another hybrid of EMAS known as a \emph{co-evolutionary multi-agent system with co-evolving species (nCoEMAS)} \cite{drezewski2003}. The main idea behind the CoEMAS model is to introduce a multiplicity of species and, thus, their co-evolution. In particular, this approach introduces different types of computational agents that are called ``niches.'' Individual agents live in so-called ``niches'' within which they can interact with other agents. Through this niching mechanism, it is possible to divide a whole population into subpopulations (species) that are located in the zones of particular local maxima in a whole solution area. The evolution of the agents takes place within individual niches; however, it is also possible to migrate between them if an agent has an adequate amount of resources. This model assumes the merging of niches if their centers are around a given maximum and the creation of new ones if a given agent has not found a suitable niche for itself to which to migrate.

Another important hybridization of EMAS is related to the use of memetic search algorithms \cite{KORCZYNSKI2017112}. Local search algorithms can gradually bring population units closer to local extremes (in this way, enhancing their genotypes). In the EMAS model, agents are autonomous entities and can conduct searches for solutions independently and freely while using the available resources. Thus, a local search could be used more often than usual (during evaluation or reproduction). A local search results in different solutions (each solution is evaluated), and the best one is selected to replace the genotype of the individual. The conducted research shows that a local search can provide a significant improvement in the obtained results in a shorter amount of time and has further potential for development. On the other hand, the application of memetic search algorithms during an agent's lifetime does not result in improvements in the obtained results. An important aspect of the work is the applied caching mechanism in the search context, which resulted in a significant increase in the number of possible fitness-function evaluations; this clearly allows for the exploration of more-difficult multi-dimensional problems.

Interesting research on EMAS is presented in \cite{Byrski2017ExtendingEO} in which several hybridizations of EDA-type and EMAS algorithms are evaluated. EDA stands for the estimation of distribution algorithms, which are universal metaheuristics that are built on evolutionary algorithms. One of the successful EDA-type algorithms is $COMMA_{op}$, which was proposed in \cite{olivier2013}. This algorithm includes exploration and exploitation phases by using a population of agents and assigning them different roles. The population of the agents evolves mostly by mutations that are governed by specific rules that are based on geometric inspirations. The idea of this research was to extend it with mechanisms of population decomposition, cloning, crossover, and death (all inspired by the EMAS approach). The introduced modifications proved successful and produced promising optimized results as compared to the original version of the $COMMA_{op}$ algorithm. In addition, the modified algorithms seem more reliable, as the dispersion of the results is significantly lower.

% \color{red}
% \textbf{Jacek Dajda}

% DO kazdej hybrydy doslownie kilka zdan na czym polega i cytowanie artykulow. Pierwsze trzy na podstawie Knowledge Engineering Review (ale cytowania trzeba dodac rowniez z cytowanych tamze prac)
% \begin{itemize}
%     \item Immunologia (byrski)
%     \item Elityzm i wielokryterialna optymalizacja (Siwik)
%     \item Niszowanie i specjacja, wiele płci (Drezewski).
%     \item Algorytmy memetyczne (Byrski/Korczynski Buffered local search for efficient memetic agent-based continuous optimization / Wojciech KORCZYŃSKI, Aleksander BYRSKI, Marek KISIEL-DOROHINICKI // Journal of Computational Science).
%     \item Estimation Distribution Algorithms and EMAS
%     (Extending estimation of distribution algorithms with agent-based computing inspirations / Aleksander BYRSKI, Marek KISIEL-DOROHINICKI, Norbert Tusiński // W: Transactions on Computational Collective Intelligence XXVII / eds. Ngoc Thanh Nguyen, RyP)
% \end{itemize}
% Jacku napisz ile mozesz, reszte Ci pomoge, a w razie czego uzupelnie.

% \color{black}

\section{EMAS hybridized with classic metaheuristics}
\begin{figure}
\centering
\includegraphics[scale=.2]{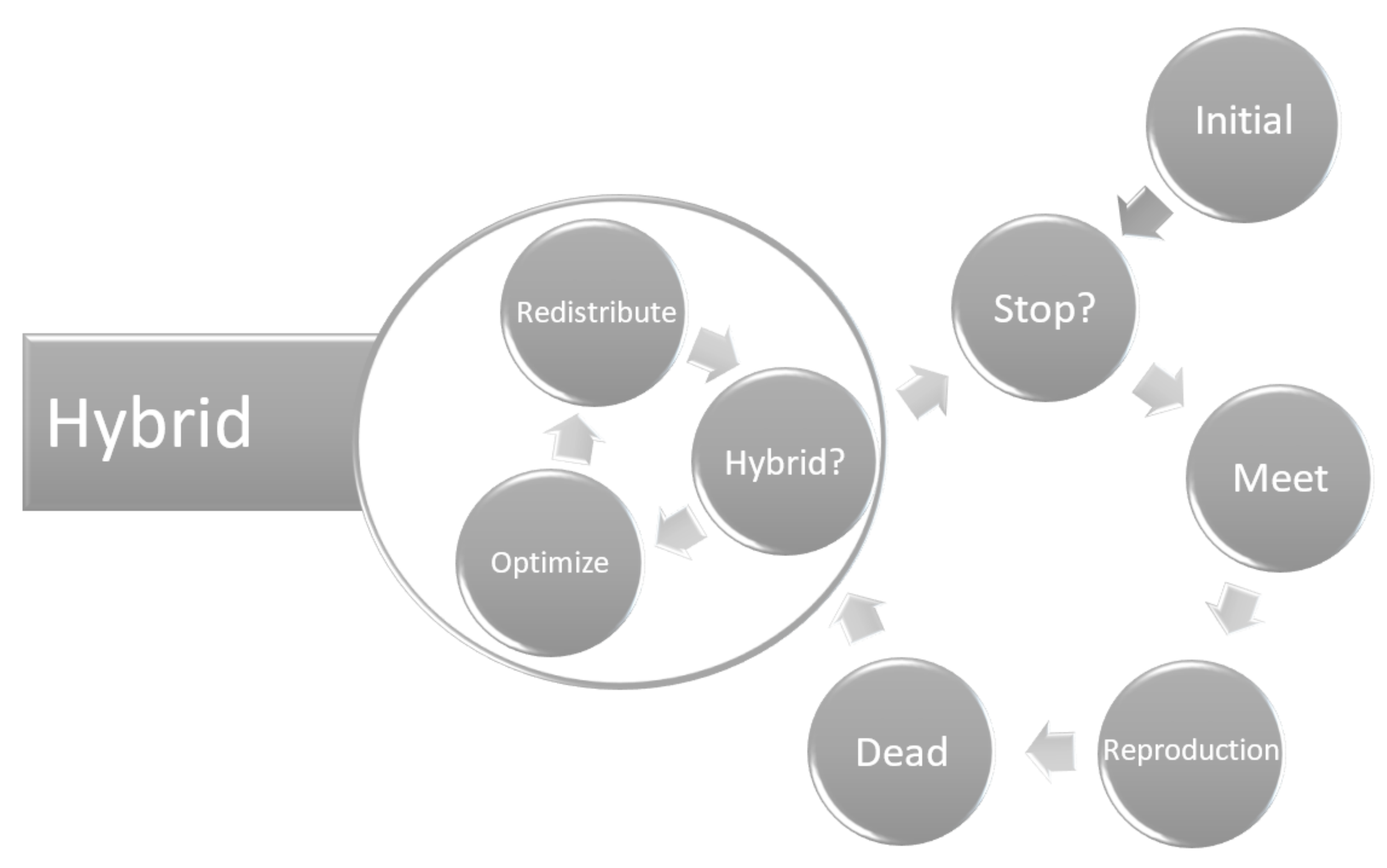}
\caption{Diagram of hybrid evolutionary multi-agent systems}
\label{rys:hemas}
\end{figure}

A hybrid evolutionary multi-agent system (HEMAS) is an algorithm that is based on EMAS that assumes an additional hybridization step. The mentioned hybridization step (shown in Figure \ref{rys:hemas}) incorporates one of the available metaheuristics into the basic EMAS algorithm. 
As HEMAS and EMAS have the same problem representation, the choice of metaheuristics is arbitrary. Should the representation be different, additional translation techniques would be required. %(see, e.g. \cite{transgen}).
The idea of hybridizing EMAS consists of running a regular computing algorithm until certain conditions are met; then, the following three phases of the algorithm are realized:
\begin{enumerate}
    \item Optimization condition: a number of different rules are available, which leads to the firing of the hybridization sequence. The rules are individual (evaluated by the agents) and may be based on dropping each agent's energy below a certain threshold after observing that the agent loses too many fights or that its offspring die very early.
    These conditions are checked globally for a certain number of algorithm steps. Those agents that are willing to participate in possible hybridizations are chosen, and the hybrid steps are executed.  
    However, failure to meet any of the conditions results in a failure to run the hybrid step. If neither of the algorithms are satisfied, this entire step is skipped. Actually, each of the used hybrid algorithms require that at least a certain (parameterized) number of agents are willing to participate. Rules can also be global -- not agent-based. In this case, a whole population is evaluated; when a rule is satisfied, the hybridizing algorithm is run for the whole agent population of agents.
    \item Running optimization algorithm:
    At this stage, the hybrid algorithms are run while incorporating the willing agents' solutions as starting ones. There is no limit to the number of algorithms or the lengths of their operations; however, it is worth choosing the algorithm that will support EMAS instead of replace it. Those algorithms can proceed in any possible way; however, they should share the same representation of the problem.
    After finishing the hybrid algorithm, the solutions are changed; therefore, the next step becomes inevitable.
    \item The energy redistribution follows here. All of the agents who participated in the previous stage must have their energy modified (as their genotypes have surely changed). Several methods of the energy redistribution were invented; namely, proportional, ranking, and tournament (following well-known selection methods \cite{goldberg}). Based on the results that are discussed in \cite{Redistribution_Godzik_2021}, we are using a proportional redistribution operator. 
\end{enumerate}

The hybridization of EMAS is depicted in Algorithm \ref{alg:hemas}. The previously known steps (cf. the EMAS algorithm \ref{alg:emas}) are supplemented with the hybridization step, which verifies the conditions of running certain hybrid algorithms, checks whether there are enough available agents that are willing to perform these steps, and delegates the genotypes of the agents to these hybridizing algorithms. Finally, the energy is redistributed according to one of the possible schemes, and EMAS continues its run. The hybridization step is realized periodically.
\begin{algorithm}[ht]
\caption{HEMAS algorithm pseudocode}\label{alg:hemas}
\begin{algorithmic}[1]
    \Function{run}{}
        \State initProgress()
        \State createInitialPopulation()

        \While{\textbf{!}isStoppingConditionReached()}
            \State meetStep()
            \State reproStep()
            \State deadStep()
            \State hybridizationStep()
            \State updateGBest()
            \State updateProgress()
        \EndWhile
    \EndFunction
\end{algorithmic}
\end{algorithm}

Within this paper, we propose a highly autonomous computing system where the decision about hybridization is fully entrusted to the agents (hybrid steps that use a particular algorithm may be used only when a minimal number of agents decide to do so).
This paper assumes simple yet efficient ways of implementing rules that trigger hybridization; however, these can be even more complex; e.g., machine-learning models (like neural networks) may be employed and learned during many of the experiments and later used as triggers for firing up the hybridization steps.

\section{Settings of hybridization}
In the discussed setting, we have chosen two classic algorithms for hybridization with EMAS:
\begin{enumerate}
    \item Particle swarm optimization \cite{pso} is an acclaimed global optimization metaheuristic algorithm that is based on particles moving in a feasible solution space and modifying their directions and velocities by using their own knowledge and the knowledge of a whole swarm to reach the optimum. 
    \item Genetic algorithm \cite{michalewicz} is a classic global optimization metaheuristic algorithm that follows the Theory of Evolution by Charles Darwin by processing a number of individuals with encoded potential solutions to a problem. These individuals undergo selection, crossover, mutation, and evaluation phases. The used version of the algorithm follows the Michalewicz model of evolution that is embedded in a real-value search space (no encoding/decoding is necessary).
\end{enumerate}
As one can see, these algorithms are completely different; however, they work on the same type of search space (classic versions of those algorithms are set in a real-value search space). Therefore, they can support basic EMAS by helping evade conditions like a lack of diversity when one of the conditions listed below are true, for example.

The conditions that are used in the presented research are local (evaluated by an individual agent) and global (evaluated globally). These conditions are constructed in order to detect unwanted situations during the search process; the above-mentioned algorithms are used to evade these situations: 
\begin{description}
  \item[VE0] -- \textbf{V}ariety \textbf{E}qual \textbf{0}: the diversity of the solutions in a population is equal to $0.0$. The diversity itself is measured as a minimal standard deviation of particular genes that are counted for a whole population. A lack of diversity is dangerous for the whole search process and can be detected when the population sticks in a local extrema. Executing a hybridizing step is aimed at the escaping of some local extrema.
  \item[ELQ1] -- \textbf{E}nergy \textbf{L}ess \textbf{1st} \textbf{Q}uartile: the energy of an agent is lower than the first quartile of the energy of a whole population of agents. Agents with low energy are about to die; such agents may tend to update their solutions by the means of a hybrid step so that they can live longer and help explore other promising areas in a search space.
  \item[EGQ3] -- \textbf{E}nergy \textbf{G}reater \textbf{3}rd \textbf{Q}uartile: the energy of an agent is higher than the third quartile of the energy of a whole population. Agents with high energy may also tend to become quite similar; therefore, running a hybrid step may also help introduce higher diversity in the best agents in a population.
  \item[VG0.5] -- \textbf{V}ariety \textbf{G}reater \textbf{0.5}: the diversity of the solutions in a population is greater than $0.5$. Detecting high diversity can be caused by the fact that a population has a very high exploratory power and is not particularly focused on exploitation at any given moment. Running the hybrid step aims at increasing the exploitation power of a whole population to focus on a search.
\end{description}

\section{Experiment design}
All of the experiments that are presented in this paper were performed on the Prometheus supercomputer, which is hosted by Academic Computing Center Cyfronet AGH; it has 2403 TFlops, runs on Linux CentOS 7, and is part of PL-Grid\footnote{http://www.plgrid.pl/en}. 

The algorithms were implemented and run using the well-known 
jMetal (Ver. 5.6) framework \footnote{jMetal\cite{jmetal}, which is a Java-based framework
that supports the development, testing, and researching of metaheuristic algorithms that are focused on single-criterion and multi-criteria optimization problems \url{http://jmetal.github.io/jMetal/}}. The version that was used was once updated by Leszek Siwik (who implemented EMAS and used it in several research papers \cite{SIWIK2018} \url{https://bitbucket.org/lesiwik/modelowaniesymulacja2018}).

In the presented research, well-known multi-dimensional Rastrigin, Ackley, Griewank, and Sphere problems with 100, 300, 500, 1000, and 2000 dimensions were used \cite{benchmarki}.
These problems had global optima in the beginning of the Cartesian space with a fitness value that was equal to $0.0$. Statistical processing was realized by using the R system.

All of the experiments were run 30 times, and the results were processed by means of descriptive statistics (means, medians, and box plots were shown) and mathematical statistics (statistical hypothesis testing that was based on Kruskal-Wallis and Dunn's test were applied). The stopping criterion for all of the tested metaheuristics was the maximum number of fitness function evaluations (this was 100 times the number of the dimensions of a problem being tackled; for a 100-dimensional problem, it was 10,000).

The EMAS and HEMAS algorithms that underwent the experiments had the following parameters:
\begin{itemize}[noitemsep]
\item Number of agents in population: $50$;
\item Total amount of resources (energy) in population: $500$;
\item Initial amount of agent's energy: $10.0$;
\item Amount of energy passed to agent that wins meeting: $1.0$;
\item Death condition: agent that reaches $0$ energy level is removed from system;
\item Reproduction condition: agent that reaches $20.0$ energy level may reproduce;
\item Crossover operator: SBXCrossover \cite{jmetal,sbx} (distribution coefficient -- $5.0$; probability -- $1.0$);
\item Mutation operator: PolynomialMutation \cite{jmetal} (distribution coefficient -- $10.0$; probability -- $0.01$);
\item Strong mutation operator: PolynomialMutation \cite{jmetal} (distribution coefficient -- $20.0$; probability -- $1.0$) (this is used when agent cannot reproduce with another agent in order to prevent effects of inbreeding).
\end{itemize}

Hybridization parameters:
\begin{itemize}[noitemsep]
\item Condition for running hybridization algorithms: periodically checking whether any rules are true and if there are agents that are willing to participate (period is $2000$ steps of  EMAS algorithm); 
\item PSO was used as hybridization algorithm (for those tests with one, two, and three operators) -- GA was only used for three-operator version;
\item Conditions for running hybridization: VE0 (one operator), ELQ1 and EGQ3 (two operators), VG0.5, ELQ1, and EGQ3 (three operators);
\item Maximum number of hybrid optimization cycles was $3$;
\item Energy redistribution operator was proportional \cite{Redistribution_Godzik_2021}.
\end{itemize}

Parameters of PSO:
\begin{itemize}
\item Swarm size: $k$;
\item Agents utilize global knowledge of whole swarm.
\end{itemize}

Parameters of GA:
\begin{itemize}[noitemsep]
\item jMetal algorithm used: GenerationalGeneticAlgorithm \cite{jmetal} (classic genetic algorithm working in real-value space according to Michalewicz model \cite{michalewicz});
\item Selection operator BinaryTournamentSelection \cite{jmetal};
\item Mutation operator PolynomialMutation (distribution coefficient -- $20.0$; probability -- inverse length of genotype [inverse dimension]);
\item Crossover operator SBXCrossover (distribution coefficient -- $20.0$; probability -- $0.9$).
\end{itemize}

\section{Results}

In the presented experiments, we focused on comparisons between EMAS and each of its hybrid versions; however, we kept in mind that comparisons between our newly proposed algorithms and classic acclaimed ones were necessary. Indeed, we showed such a comparison in \cite{GodzikGPSBK19}, where one can find the results that indicate the better efficacy of the proposed hybrid agent-based algorithms as compared to the acclaimed differential evolution algorithm.

\subsection{HEMAS with one hybridization operator}
\begin{figure}[ht]
    \centering
    \includegraphics[width=0.98\textwidth]{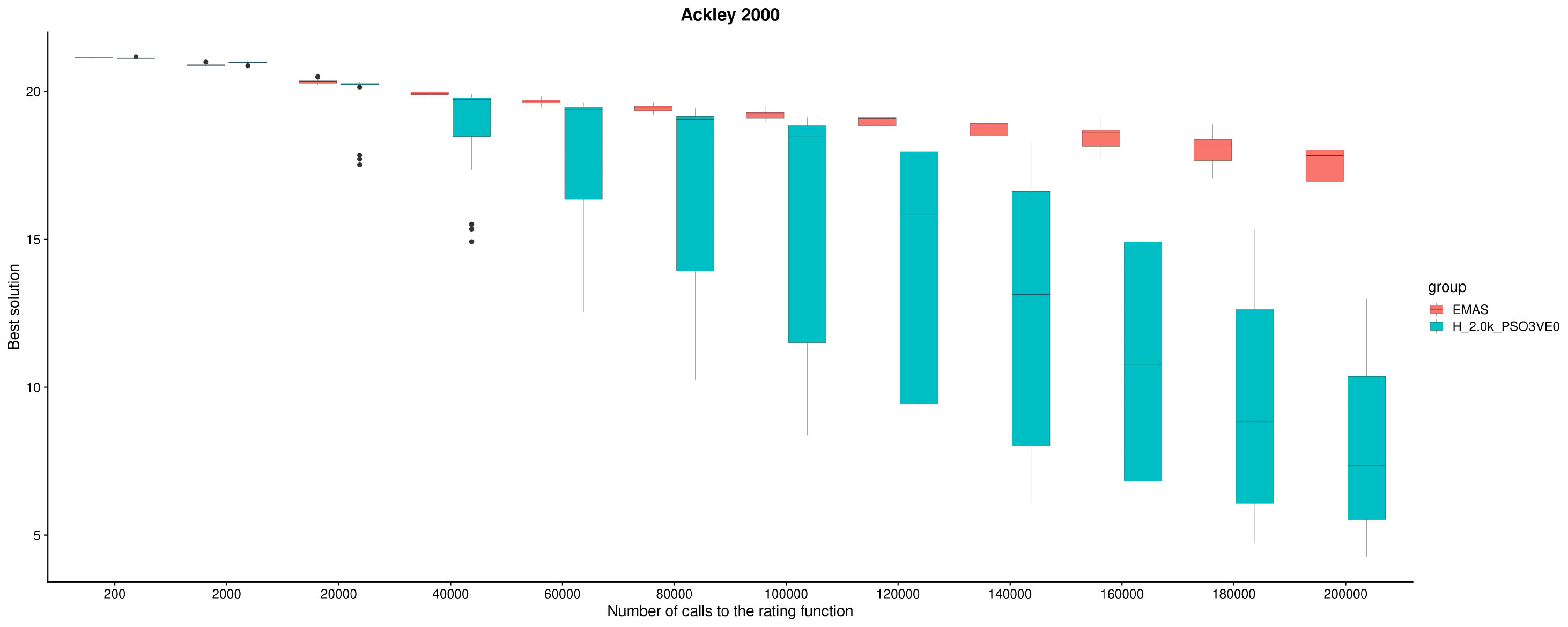}
        \caption{Best solution in current step of algorithm dependent on number of fitness evaluation: EMAS (red); HEMAS (blue); one hybridization operator (VE0 hybridization condition); Ackley in 2000 dimensions}
        \label{fig:partial_results_1}
\end{figure}

Fig. \ref{fig:partial_results_1} shows the best fitness value in the current population that is dependent on the number of fitness function calls for the Ackley benchmark (in 2000 dimensions). PSO was the hybridization operator, and it was run when the diversity of the solutions fell to $0$; therefore, it might be considered to be a last-resort algorithm for preventing premature convergence or a longer period of time when no promising results are found, for example.
HEMAS was been tested with other operators (the genetic [GA], differential evolution [DE], simulated annealing [SA], and evolution strategy [ES] algorithms) and different startup conditions (energy below level [EL], energy above level [EG], agent solution below level [SL], agent solution above level [SG], variety of solutions below level [VL], and variety of solutions above level [VG]) as well as other condition levels (both constant as 3, 5, 7, 10, 15, and 17 as well as dynamic from population as average [M], first quartile [Q1], median [Q2], and third quartile [Q3]); however, the combination described above obtained the best results.
The graph in Fig. \ref{fig:partial_results_1} shows box-and-whisker plots of the averaged fitnesses over all of the experimental runs of the algorithm. As one can see, both algorithms produced quite similar results during the first 20,000 evaluations of the fitness function; apparently, the hybrid version then began to approach the optimum, while the classic version was apparently stuck in some local extrema and was able to increase its quality much more slowly than with the hybrid version. Moreover, one can see that the hybrid version had a higher exploratory power than the classic version did because of the significantly higher dispersion of the results. To sum up, EMAS did not lose its searching potential, but HEMAS apparently searched in a more efficient and effective manner.

\begin{figure}[p]
    \centering
        % \begin{subfigure}[b]{0.32\textwidth}
        %     \includegraphics[width=\textwidth]{graphics/z1/AverageAckley 100.pdf}
        %     \caption{100D funkcja Ackley}
        %     \label{fig:z1_ackley_100}
        % \end{subfigure}
        % \begin{subfigure}[b]{0.32\textwidth}
        %     \includegraphics[width=\textwidth]{graphics/z1/AverageAckley 300.pdf}
        %     \caption{300D funkcja Ackley}
        %     \label{fig:z1_ackley_300}
        % \end{subfigure}
        % \begin{subfigure}[b]{0.32\textwidth}
        %     \includegraphics[width=\textwidth]{graphics/z1/AverageAckley 500.pdf}
        %     \caption{500D funkcja Ackley}
        %     \label{fig:z1_ackley_500}
        % \end{subfigure}
        \begin{subfigure}[b]{0.32\textwidth}
            \includegraphics[width=\textwidth]{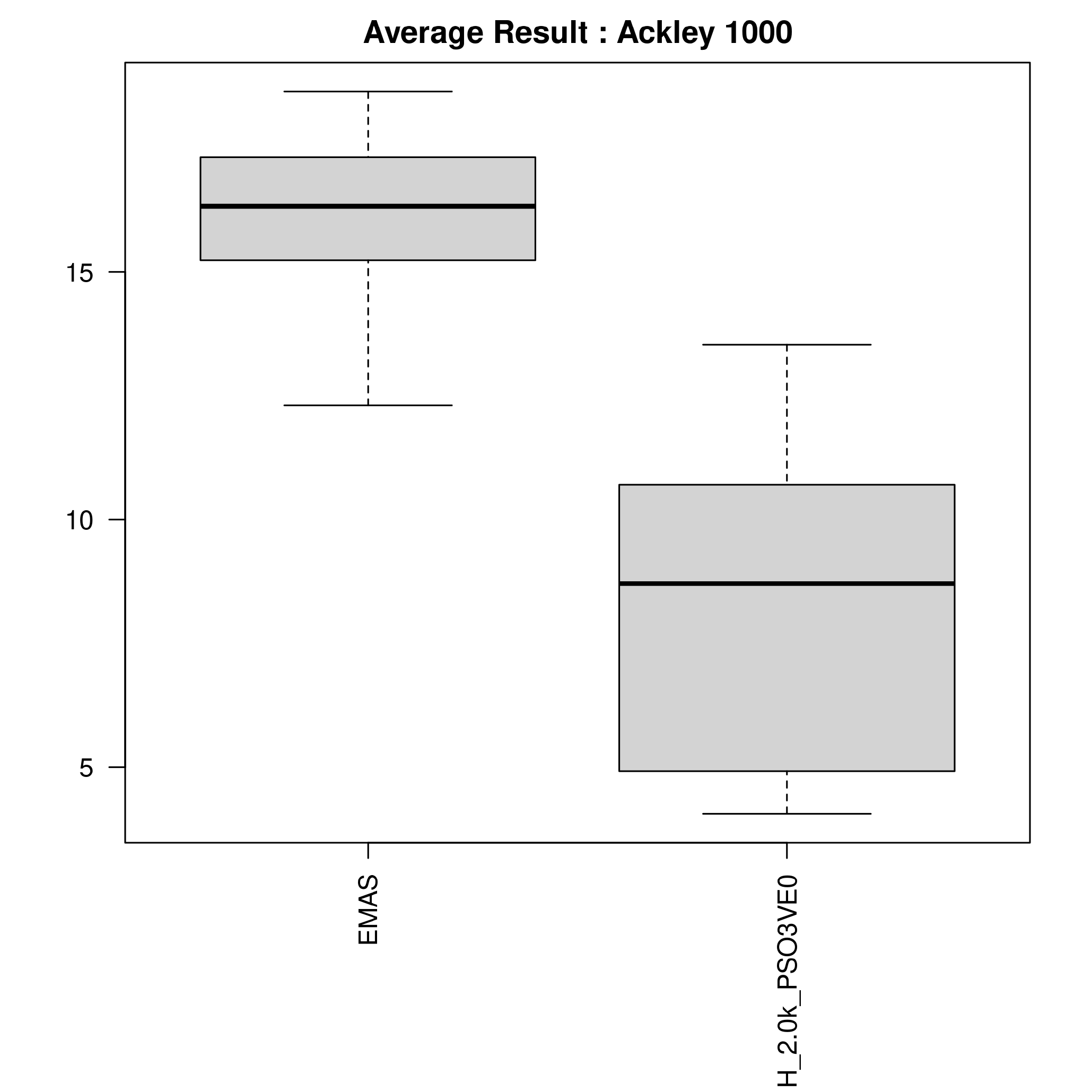}
            \caption{Ackley 1000D}
            \label{fig:z1_ackley_1000}
        \end{subfigure}
        \begin{subfigure}[b]{0.32\textwidth}
            \includegraphics[width=\textwidth]{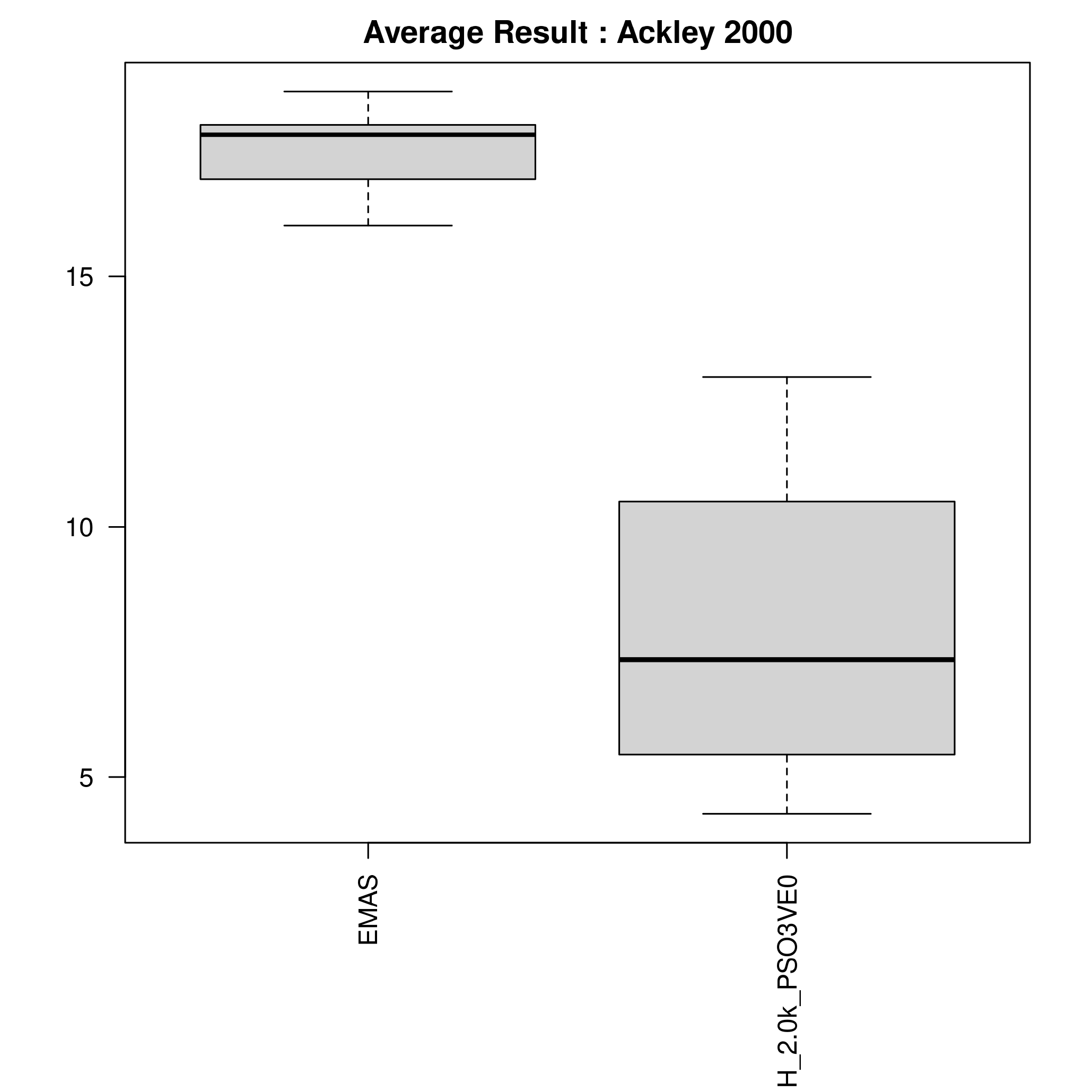}
            \caption{Ackley 2000D}
            \label{fig:z1_ackley_2000}
        \end{subfigure}
        % \begin{subfigure}[b]{0.32\textwidth}
        %     \includegraphics[width=\textwidth]{graphics/z1/AverageRastrigin 100.pdf}
        %     \caption{100D funkcja Rastrigin}
        %     \label{fig:z1_rastrigin_100}
        % \end{subfigure}
        % \begin{subfigure}[b]{0.32\textwidth}
        %     \includegraphics[width=\textwidth]{graphics/z1/AverageRastrigin 300.pdf}
        %     \caption{300D funkcja Rastrigin}
        %     \label{fig:z1_rastrigin_300}
        % \end{subfigure}
        % \begin{subfigure}[b]{0.32\textwidth}
        %     \includegraphics[width=\textwidth]{graphics/z1/AverageRastrigin 500.pdf}
        %     \caption{500D funkcja Rastrigin}
        %     \label{fig:z1_rastrigin_500}
        % \end{subfigure}
        \begin{subfigure}[b]{0.32\textwidth}
            \includegraphics[width=\textwidth]{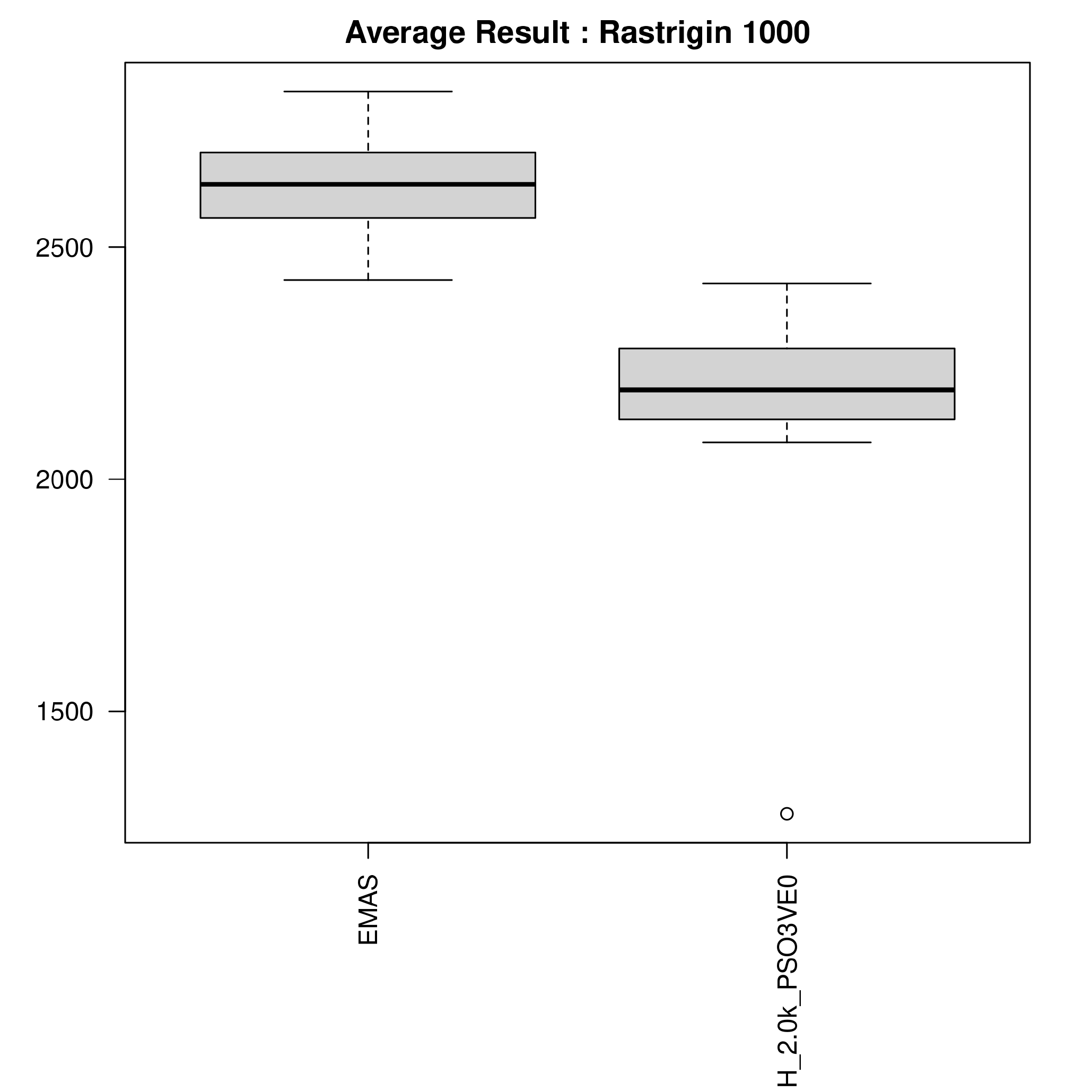}
            \caption{Rastrigin 1000D}
            \label{fig:z1_rastrigin_1000}
        \end{subfigure}
        \begin{subfigure}[b]{0.32\textwidth}
            \includegraphics[width=\textwidth]{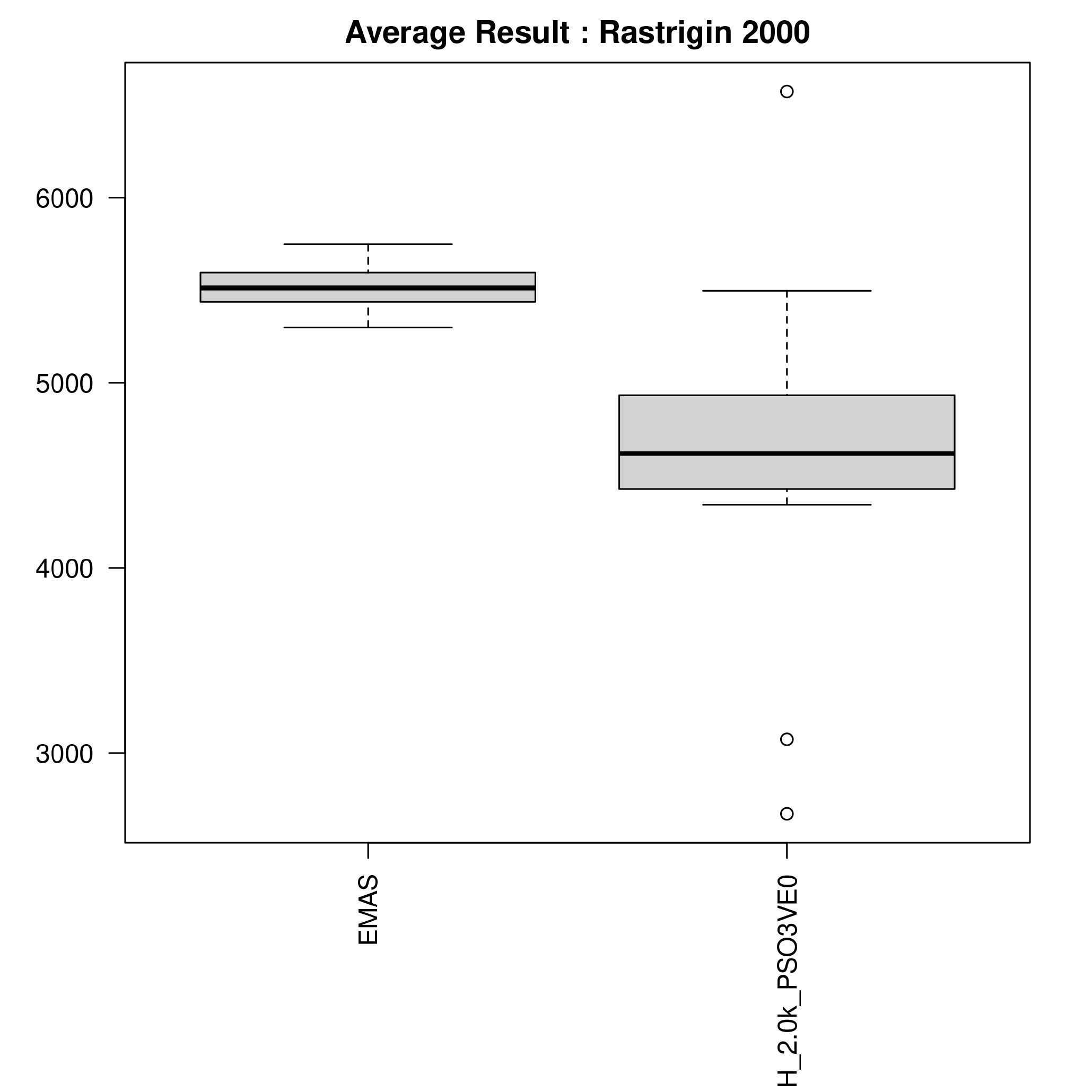}
            \caption{Rastrigin 2000D}
            \label{fig:z1_rastrigin_2000}
        \end{subfigure}
         \begin{subfigure}[b]{0.32\textwidth}
            \includegraphics[width=\textwidth]{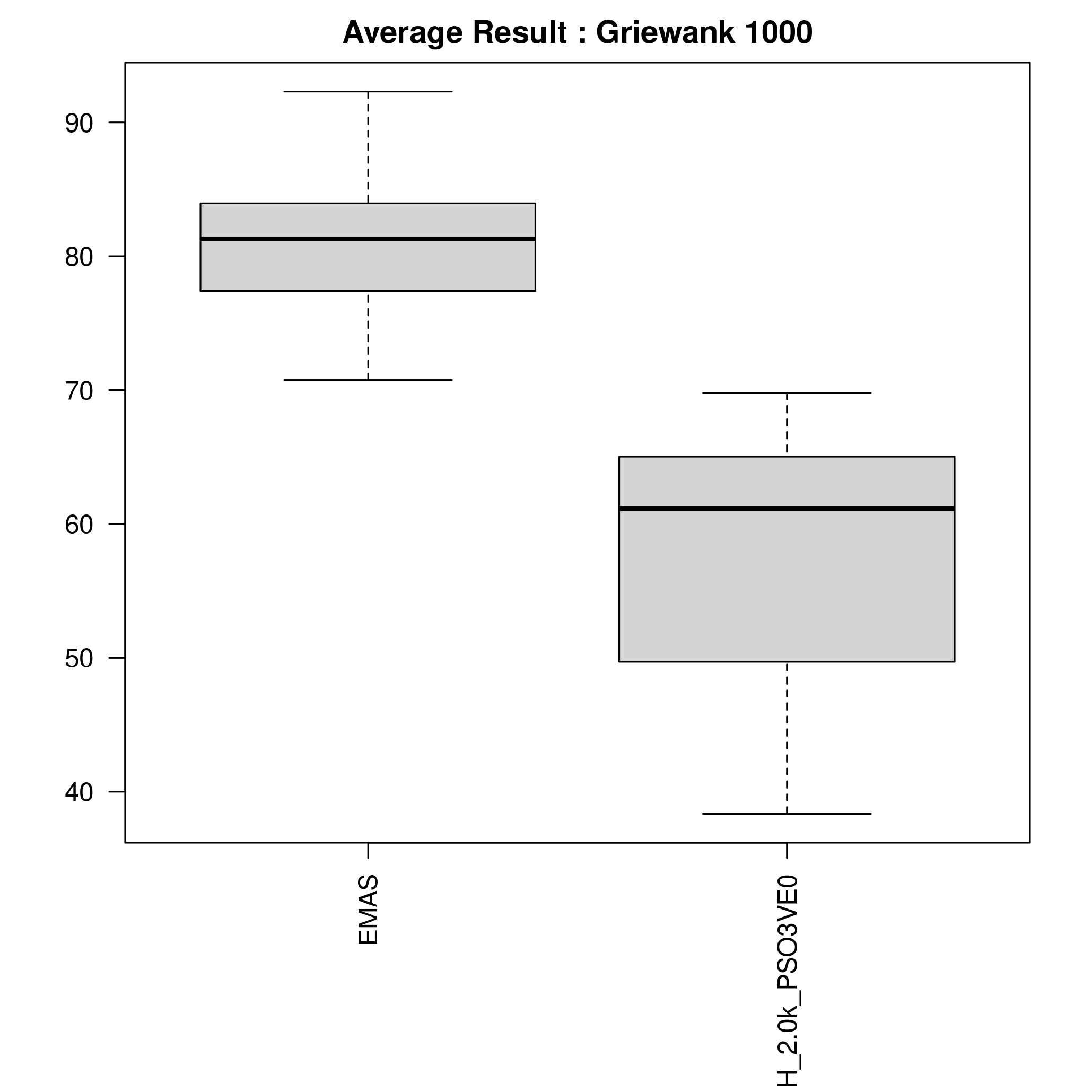}
            \caption{Griewank 1000D}
            \label{fig:z1_griewank_1000}
        \end{subfigure}
        \begin{subfigure}[b]{0.32\textwidth}
            \includegraphics[width=\textwidth]{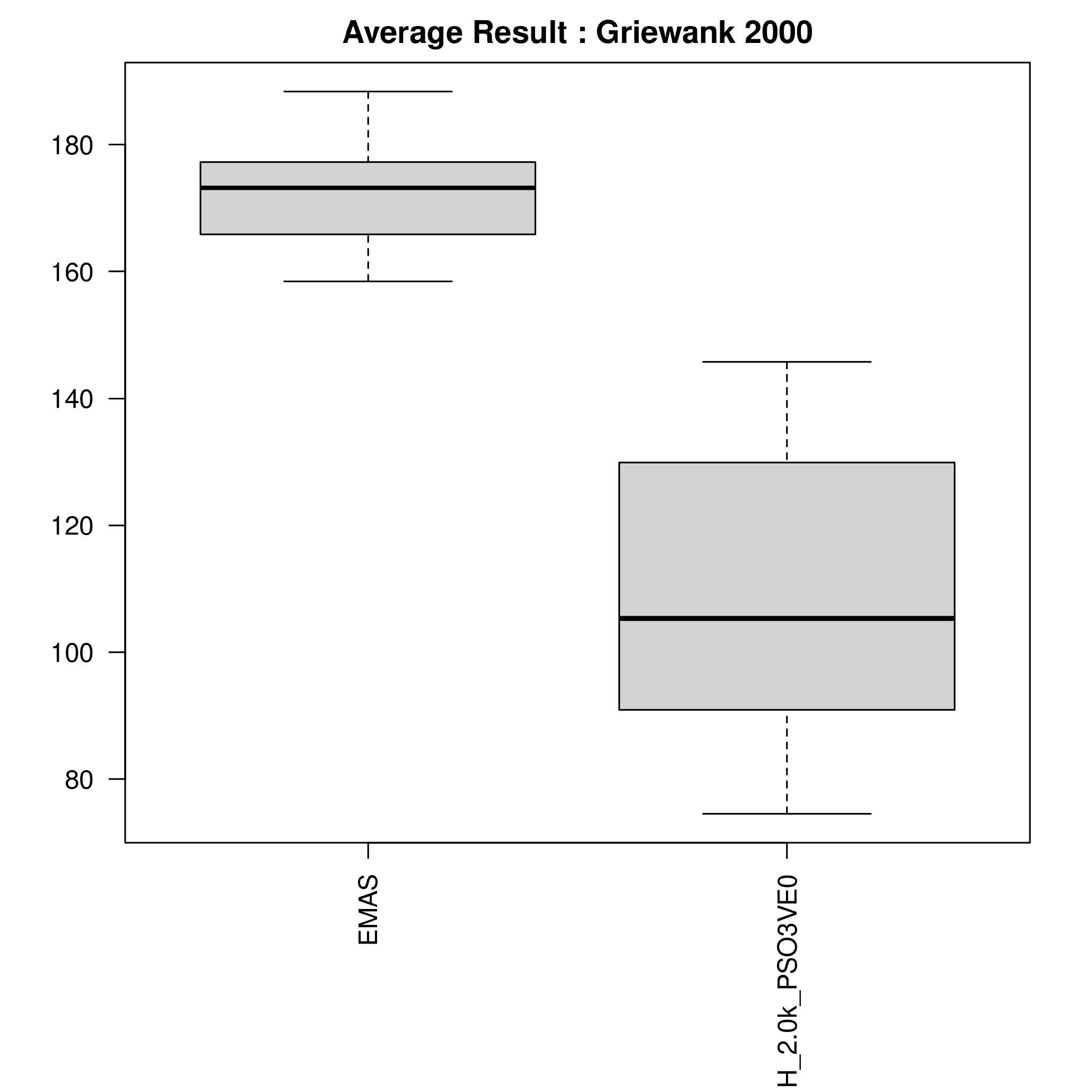}
            \caption{Griewank 2000D}
            \label{fig:z1_griewank_2000}
        \end{subfigure}
        \begin{subfigure}[b]{0.32\textwidth}
            \includegraphics[width=\textwidth]{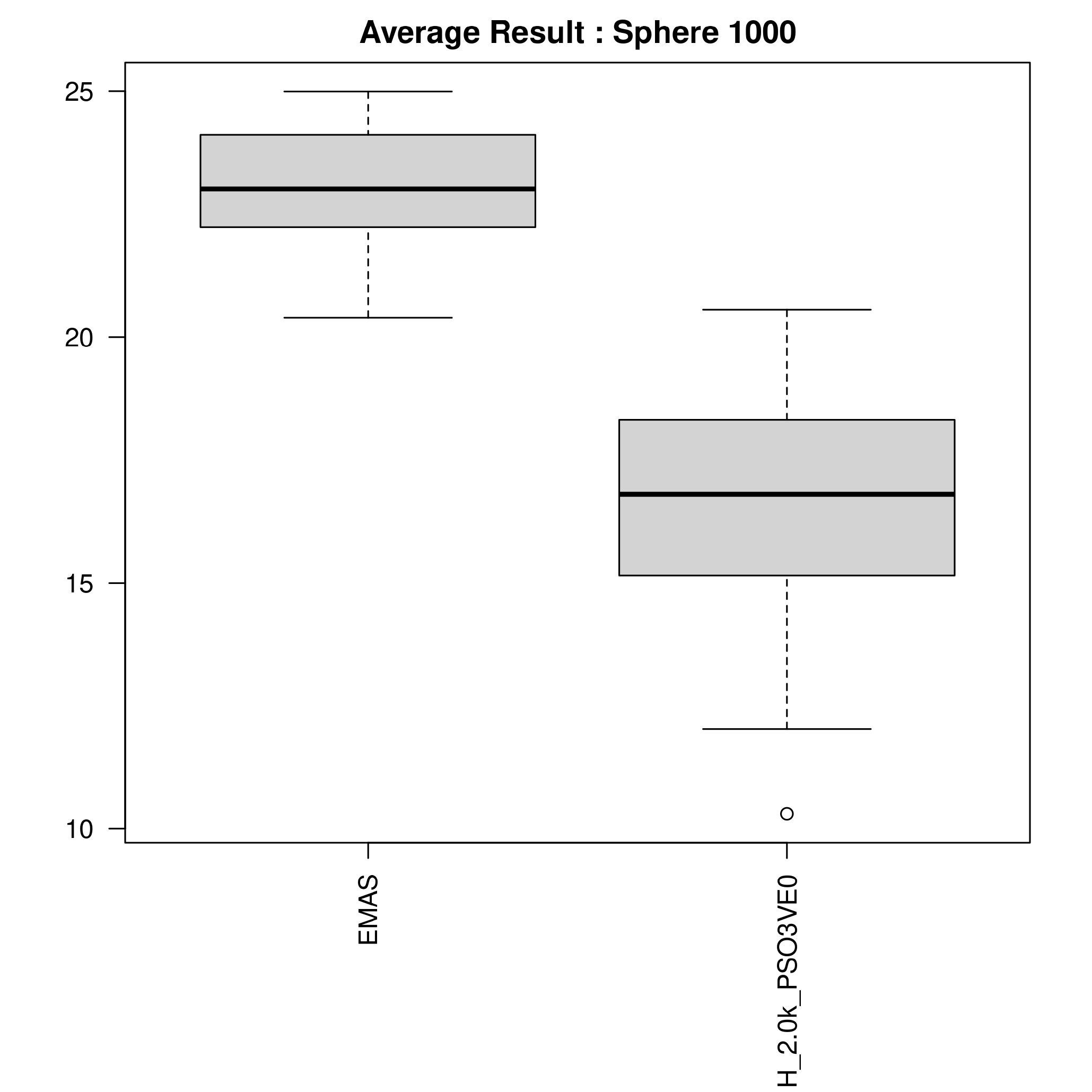}
            \caption{Sphere 1000D}
            \label{fig:z1_sphere_1000}
        \end{subfigure}
        \begin{subfigure}[b]{0.32\textwidth}
            \includegraphics[width=\textwidth]{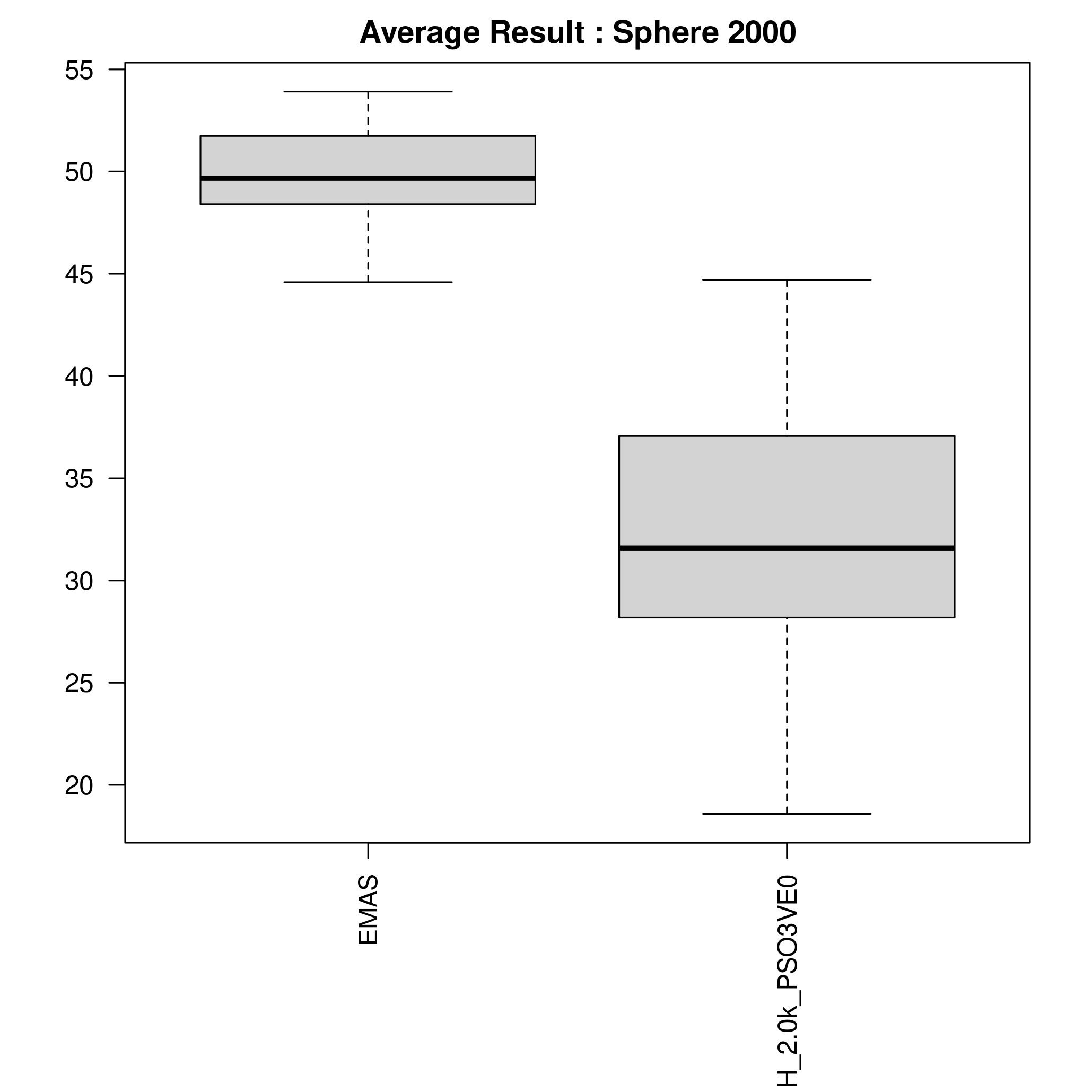}
            \caption{Sphere 2000D}
            \label{fig:z1_sphere_2000}
        \end{subfigure}
        
    \caption{Final fitnesses for selected benchmarks that were obtained by EMAS and HEMAS with PSO hybridization operator (Rule VE0)}
    \label{fig:final_z1_1}
\end{figure}

In order to have a broader view of the efficacy and efficiency of HEMAS, let us take a look at Fig. \ref{fig:final_z1_1}; this shows the final results of EMAS and HEMAS that were computed for the most difficult benchmarks that were tackled. In all of the observed cases, the final solutions that were obtained for the hybrid version turned out to be visibly better than those of the classic one. Moreover, the observed whiskers did not overlap, which allowed us to predict that the observed differences were statistically significant. One should note that the Rastrigin benchmark (Figs. \ref{fig:z1_rastrigin_1000} and \ref{fig:z1_rastrigin_2000}) turned out to be the most difficult for both the tested hybrid and the original algorithm; however, the algorithm that was run for the other benchmarks (Ackley, Sphere, and Griewank) was able to finish the search much closer to the global optimum.
    
These observations can be confirmed when looking at Table \ref{tab:Results1}, where the EMAS and HEMAS final results can be found (along with the necessary descriptive statistic factors). Both algorithms were able to approach the global extrema (which was equal to $0$ for all of the problems tackled); however, HEMAS seemed to be closer. For each test case, the means, medians, minimum values, and maximum values (except for Rastrigin 2000) of the results were better for HEMAS with one optimization operator. EMAS had better standard deviations in most cases, which may result in faster stalling at local minima.

\begingroup
\renewcommand{\arraystretch}{0.75}
\begin{table}
\centering
    \caption{Results of EMAS and HEMAS with one optimizer operator for tested problems (better values shown as \textbf{bold}) \label{tab:Results1}}
    \begin{tabular}{llllll}
        \toprule
        \multicolumn{6}{c}{EMAS } \\
            & {Mean} & {Median} & {SD} & {Minimum} & {Maximum} \\
        \midrule
        \textbf{Ackley 100}     & 5.926268 & 5.597584 & 0.963522 & 4.520213 & 8.211922 \\ 
        \textbf{Ackley 300}     & 9.762755 & 9.279761 & 2.326365 & 6.527268 & 16.21522 \\ 
        \textbf{Ackley 500}     & 12.34542 & 12.0716  & \textbf{2.203618} & 8.912715 & 16.84377 \\ 
        \textbf{Ackley 1000}    & 16.23966 & 16.33226 & \textbf{1.438965} & 12.30445 & 18.64724 \\ 
        \textbf{Ackley 2000}    & 17.54132 & 17.83075 & \textbf{0.749609} & 16.01603 & 18.6924  \\ 
        \textbf{Griewank 100}   & 6.17848  & 6.161992 & \textbf{0.641377} & 5.078244 & 7.489971 \\ 
        \textbf{Griewank 300}   & 20.0875  & 20.27968 & \textbf{1.631232} & 17.23735 & 22.83947 \\ 
        \textbf{Griewank 500}   & 36.71405 & 37.00883 & \textbf{3.065954} & 29.47909 & 44.58615 \\ 
        \textbf{Griewank 1000}  & 80.79249 & 81.28187 & \textbf{4.920894} & 70.7338  & 92.29649 \\ 
        \textbf{Griewank 2000}  & 172.3125 & 173.1885 & \textbf{7.167818} & 158.4439 & 188.3559 \\ 
        \textbf{Rastrigin 100}  & 201.738  & 204.3074 & 25.60595 & 148.4729 & 258.7543 \\ 
        \textbf{Rastrigin 300}  & 706.7516 & 696.8028 & \textbf{41.35456} & 600.2623 & 804.0281 \\ 
        \textbf{Rastrigin 500}  & 1240.539 & 1245.277 & \textbf{59.37643} & 1129.241 & 1356.073 \\ 
        \textbf{Rastrigin 1000} & 2629.001 & 2635.377 & \textbf{95.09159} & 2429.206 & 2835.257 \\ 
        \textbf{Rastrigin 2000} & 5515.959 & 5513.418 & \textbf{121.274}  & 5298.585 & \textbf{5747.834} \\ 
        \textbf{Sphere 100}     & 1.544541 & 1.578112 & 0.257854 & 1.110325 & 1.980352 \\ 
        \textbf{Sphere 300}     & 5.853364 & 5.896428 & 0.663037 & 4.728222 & 7.31008  \\ 
        \textbf{Sphere 500}     & 10.45616 & 10.46227 & \textbf{0.729824} & 8.920684 & 12.24955 \\ 
        \textbf{Sphere 1000}    & 23.03572 & 23.00944 & \textbf{1.169066} & 20.39787 & 24.99274 \\ 
        \textbf{Sphere 2000}    & 49.90044 & 49.67515 & \textbf{2.442108} & 44.59093 & 53.91276 \\ 
        \toprule
        \multicolumn{6}{c}{HEMAS with one optimizer operator } \\
            & {Mean} & {Median} & {SD} & {Minimum} & {Maximum} \\
        \midrule
        \textbf{Ackley 100}     & \textbf{5.56755}  & \textbf{5.360173} & \textbf{0.797002} & \textbf{4.399874} & \textbf{8.000223} \\
        \textbf{Ackley 300}     & \textbf{7.682239} & \textbf{7.464139} & \textbf{1.442116} & \textbf{4.073859} & \textbf{10.89752} \\
        \textbf{Ackley 500}     & \textbf{8.584816} & \textbf{8.240185} & 2.902486 & \textbf{4.835378} & \textbf{16.48923} \\
        \textbf{Ackley 1000}    & \textbf{8.073076} & \textbf{8.706925} & 2.914813 & \textbf{4.05603}  & \textbf{13.53742} \\
        \textbf{Ackley 2000}    & \textbf{7.902813} & \textbf{7.349659} & 2.707326 & \textbf{4.26931}  & \textbf{12.98765} \\
        \textbf{Griewank 100}   & \textbf{5.565565} & \textbf{5.617452} & 0.702329 & \textbf{4.369381} & \textbf{7.489822} \\
        \textbf{Griewank 300}   & \textbf{17.94343} & \textbf{17.54779} & 1.711432 & \textbf{15.57559} & \textbf{22.51006} \\
        \textbf{Griewank 500}   & \textbf{29.3409}  & \textbf{29.9368}  & 3.306269 & \textbf{20.48791} & \textbf{35.41683} \\
        \textbf{Griewank 1000}  & \textbf{58.11122} & \textbf{61.12984} & 9.459287 & \textbf{38.34293} & \textbf{69.77153} \\
        \textbf{Griewank 2000}  & \textbf{110.447}  & \textbf{105.3043} & 22.39588 & \textbf{74.54867} & \textbf{145.7659} \\
        \textbf{Rastrigin 100}  & \textbf{184.4665} & \textbf{190.3774} & \textbf{24.2931}  & \textbf{121.2793} & \textbf{220.9861} \\
        \textbf{Rastrigin 300}  & \textbf{610.8486} & \textbf{613.6308} & 47.9938  & \textbf{522.675}  & \textbf{696.3187} \\
        \textbf{Rastrigin 500}  & \textbf{1086.845} & \textbf{1085.976} & 80.06016 & \textbf{938.6453} & \textbf{1298.339} \\
        \textbf{Rastrigin 1000} & \textbf{2180.257} & \textbf{2192.65}  & 194.073  & \textbf{1279.293} & \textbf{2421.913} \\
        \textbf{Rastrigin 2000} & \textbf{4662.964} & \textbf{4618.283} & 656.3485 & \textbf{2672.92}  & 6573.332 \\
        \textbf{Sphere 100}     & \textbf{1.240428} & \textbf{1.228573} & \textbf{0.229846} & \textbf{0.826595} & \textbf{1.683883} \\
        \textbf{Sphere 300}     & \textbf{4.667651} & \textbf{4.800594} & \textbf{0.528825} & \textbf{3.685149} & \textbf{5.613926} \\
        \textbf{Sphere 500}     & \textbf{7.98475}  & \textbf{8.392535} & 1.088776 & \textbf{5.31472}  & \textbf{9.617134} \\
        \textbf{Sphere 1000}    & \textbf{16.36833} & \textbf{16.80346} & 2.449157 & \textbf{10.30567} & \textbf{20.5579}  \\
        \textbf{Sphere 2000}    & \textbf{32.50936} & \textbf{31.58886} & 6.281789 & \textbf{18.58552} & \textbf{44.70344} \\
        \bottomrule
    \end{tabular}
	
\end{table}
\endgroup

\subsection{HEMAS with two hybridization operators}
\begin{figure}[ht]
    \centering
    \includegraphics[width=0.98\textwidth]{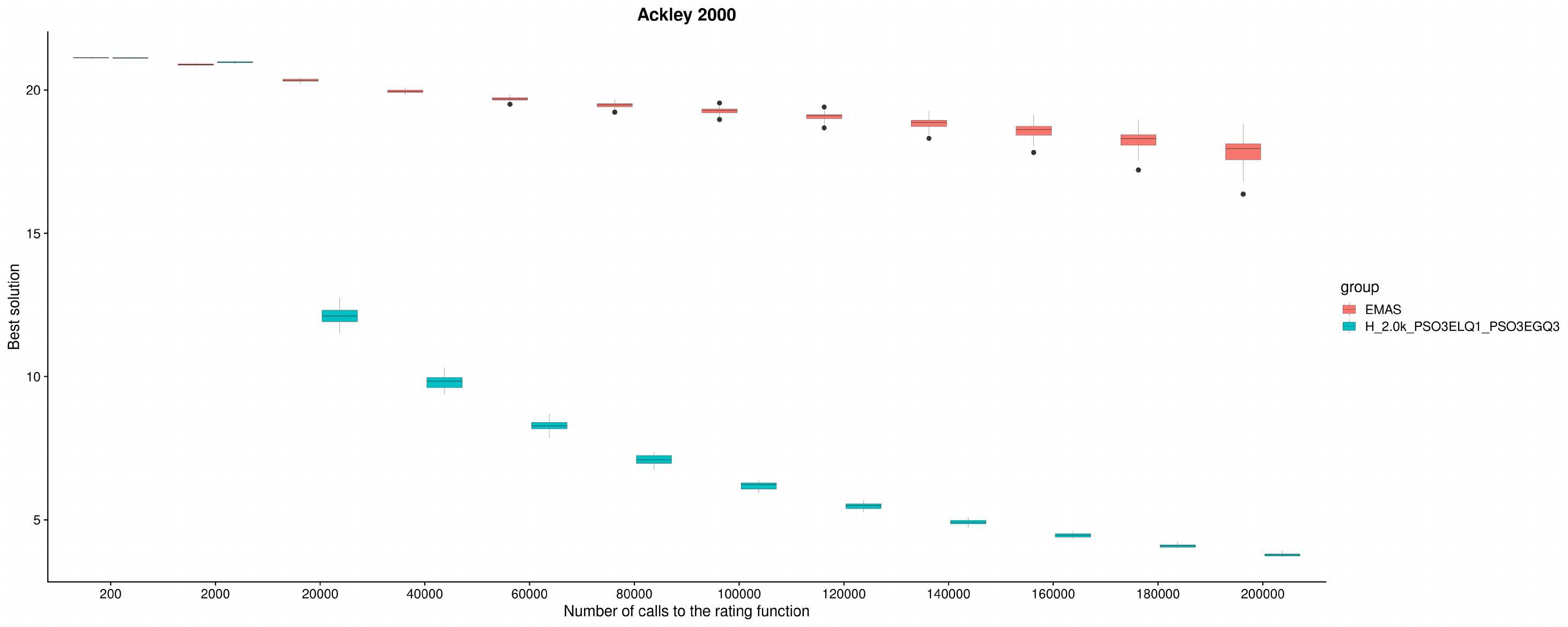}
        \caption{Best fitnesses for EMAS and HEMAS with two PSO hybridization operators (Rules  ELQ1 and  EGQ3) for  Ackley problem in 2000 dimensions (depending on number of fitness function calls)}
        \label{fig:partial_results_2}
\end{figure}

In Fig. \ref{fig:partial_results_2}, the best solution for EMAS and HEMAS is shown when two PSO hybridization operators were applied. These were executed for those agents that had energy levels that were lower than the first quartile of the energy in the system as well as for those agents that had energy levels that were higher than the third quartile of the energy in the system. This combination was chosen as the best after testing all of the possible combinations of operator pairs (PSO, DE, GA, SA, and ES) and conditions (EL3 + VE0, EL3 + VG0.5, EL3 + VG1, EG17 + VE0, EG17 + VG0.5, EG17 + VG1, EL3 + EG17, ELQ1 + EGQ3, and SLQ1 + SGQ3). For the observed 2000D Ackley problem (Fig. \ref{fig:partial_results_2}), HEMAS started to beat EMAS very quickly (starting with $20,000$ evaluations of the fitness function). EMAS apparently got stuck in a local optimum, and HEMAS retained its explorative power thanks to the application of hybrid operators. It is to note that the dispersion of the results was lower than it was in the case of one hybridization operator -- the whole experiment was repeatable (the dispersion of the observed results was reasonable). The rationale for applying these two rules for running PSO was to rescue the worst solutions from imminent death and to improve the best solutions in order to get even closer to the optimum.

\begin{figure}[p]
    \centering
        % \begin{subfigure}[b]{0.32\textwidth}
        %     \includegraphics[width=\textwidth]{graphics/z2/AverageAckley 100.pdf}
        %     \caption{100D funkcja Ackley}
        %     \label{fig:z2_ackley_100}
        % \end{subfigure}
        % \begin{subfigure}[b]{0.32\textwidth}
        %     \includegraphics[width=\textwidth]{graphics/z2/AverageAckley 300.pdf}
        %     \caption{300D funkcja Ackley}
        %     \label{fig:z2_ackley_300}
        % \end{subfigure}
        % \begin{subfigure}[b]{0.32\textwidth}
        %     \includegraphics[width=\textwidth]{graphics/z2/AverageAckley 500.pdf}
        %     \caption{500D funkcja Ackley}
        %     \label{fig:z2_ackley_500}
        % \end{subfigure}
        \begin{subfigure}[b]{0.32\textwidth}
            \includegraphics[width=\textwidth]{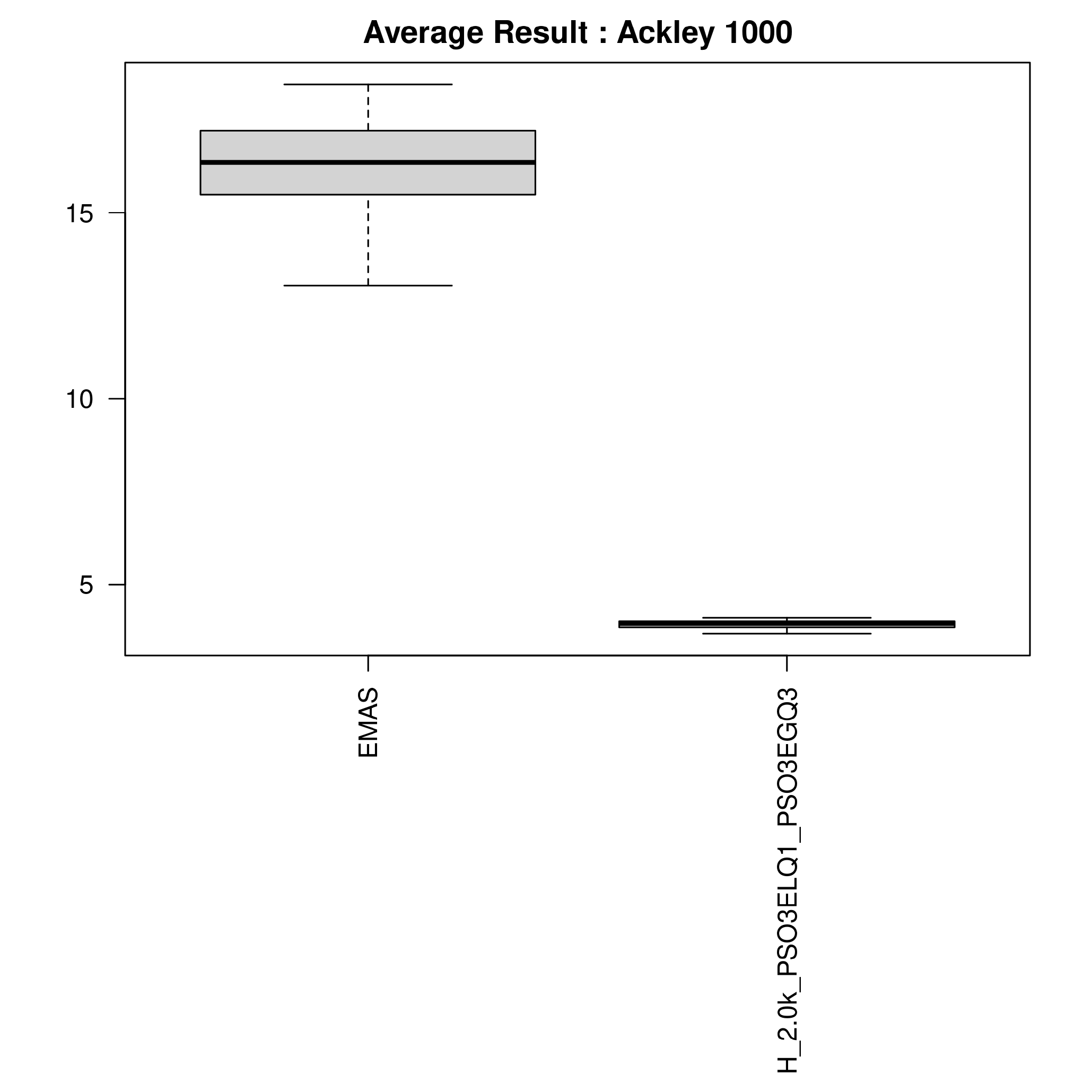}
            \caption{Ackley 1000D}
            \label{fig:z2_ackley_1000}
        \end{subfigure}
        \begin{subfigure}[b]{0.32\textwidth}
            \includegraphics[width=\textwidth]{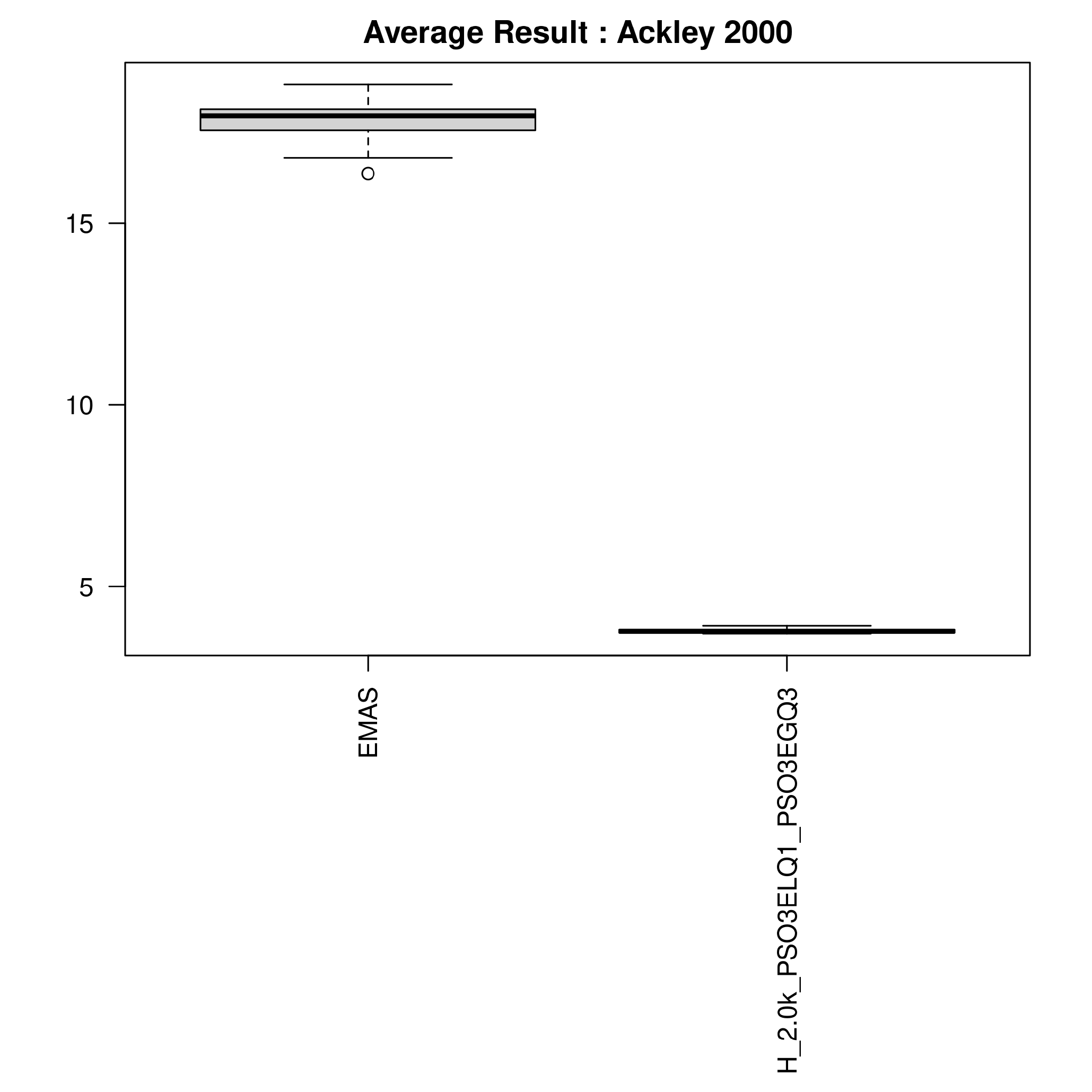}
            \caption{Ackley 2000D}
            \label{fig:z2_ackley_2000}
        \end{subfigure}
        % \begin{subfigure}[b]{0.32\textwidth}
        %     \includegraphics[width=\textwidth]{graphics/z2/AverageRastrigin 100.pdf}
        %     \caption{100D funkcja Rastrigin}
        %     \label{fig:z2_rastrigin_100}
        % \end{subfigure}
        % \begin{subfigure}[b]{0.32\textwidth}
        %     \includegraphics[width=\textwidth]{graphics/z2/AverageRastrigin 300.pdf}
        %     \caption{300D funkcja Rastrigin}
        %     \label{fig:z2_rastrigin_300}
        % \end{subfigure}
        % \begin{subfigure}[b]{0.32\textwidth}
        %     \includegraphics[width=\textwidth]{graphics/z2/AverageRastrigin 500.pdf}
        %     \caption{500D funkcja Rastrigin}
        %     \label{fig:z2_rastrigin_500}
        % \end{subfigure}
        \begin{subfigure}[b]{0.32\textwidth}
            \includegraphics[width=\textwidth]{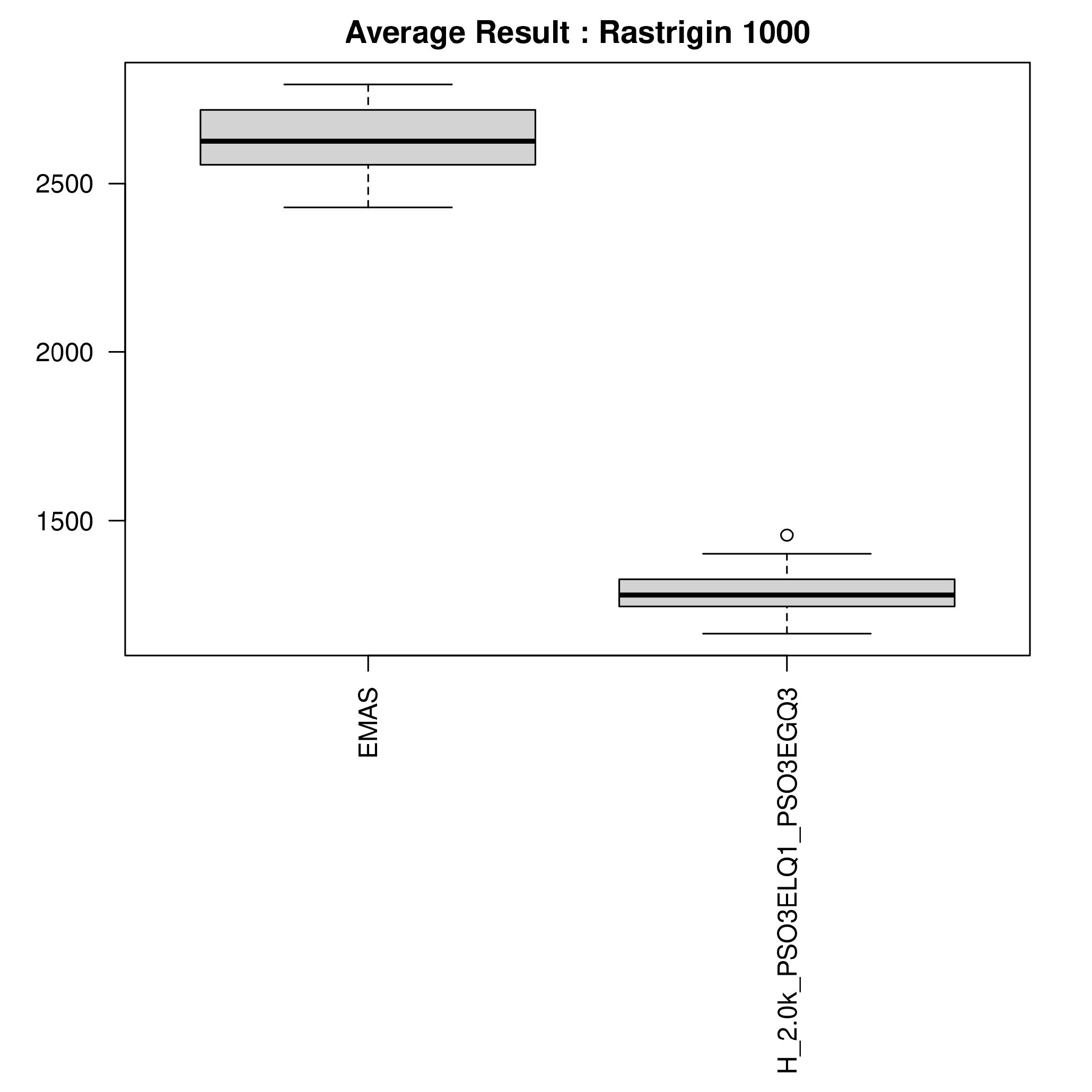}
            \caption{Rastrigin 1000D}
            \label{fig:z2_rastrigin_1000}
        \end{subfigure}
         \begin{subfigure}[b]{0.32\textwidth}
            \includegraphics[width=\textwidth]{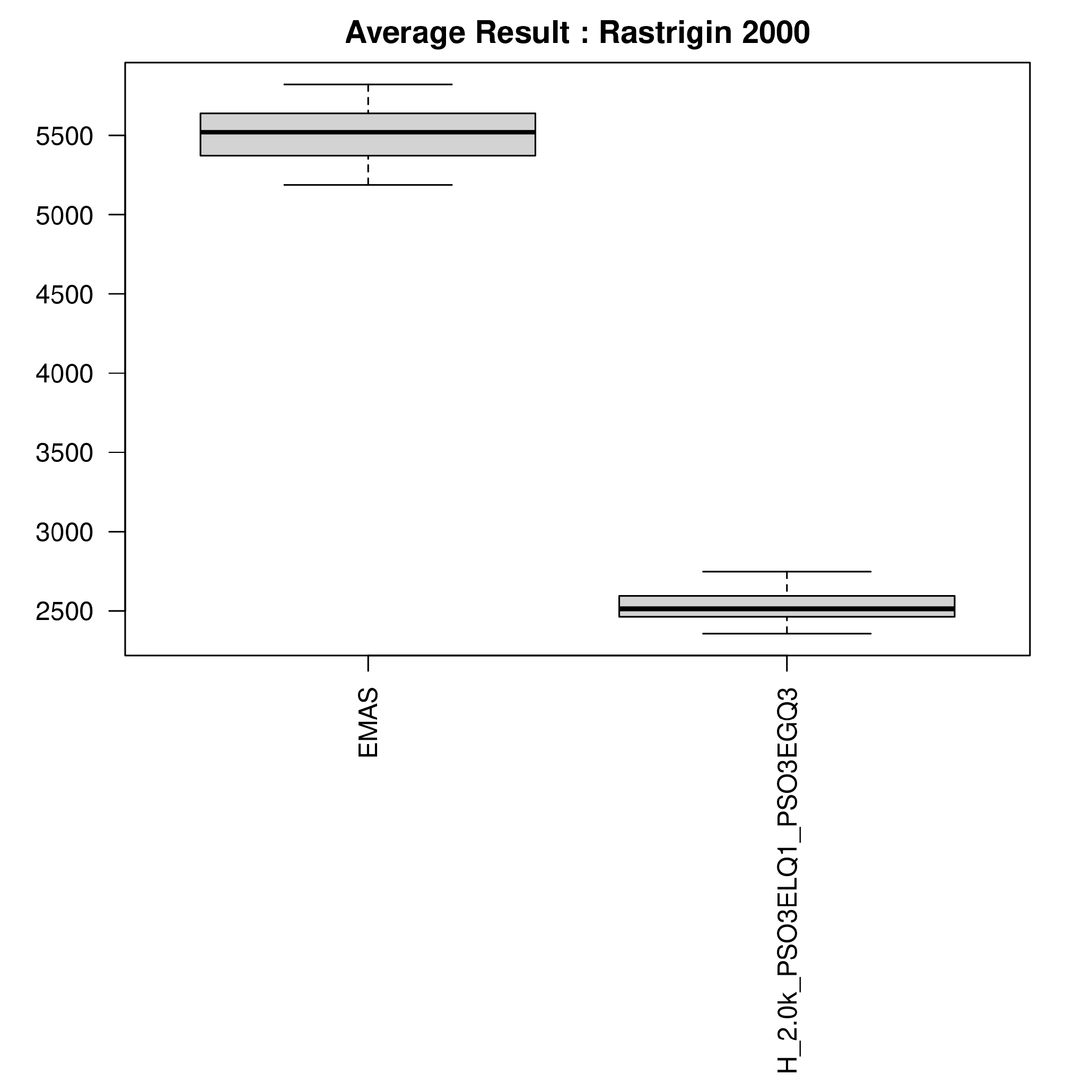}
            \caption{Rastrigin 2000D}
            \label{fig:z2_rastrigin_2000}
        \end{subfigure}
         \begin{subfigure}[b]{0.32\textwidth}
            \includegraphics[width=\textwidth]{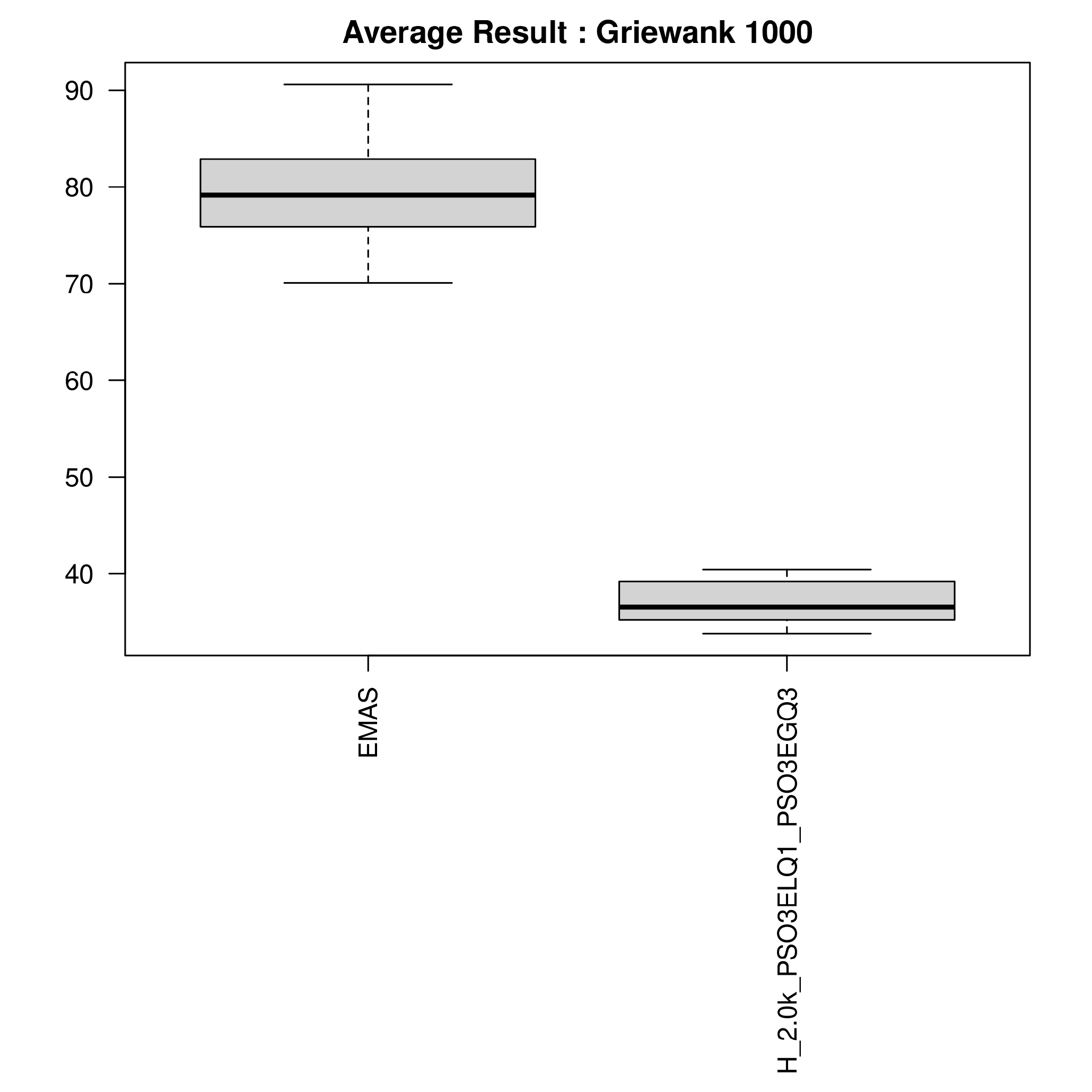}
            \caption{Griewank 1000D}
            \label{fig:z2_griewank_1000}
        \end{subfigure}
        \begin{subfigure}[b]{0.32\textwidth}
            \includegraphics[width=\textwidth]{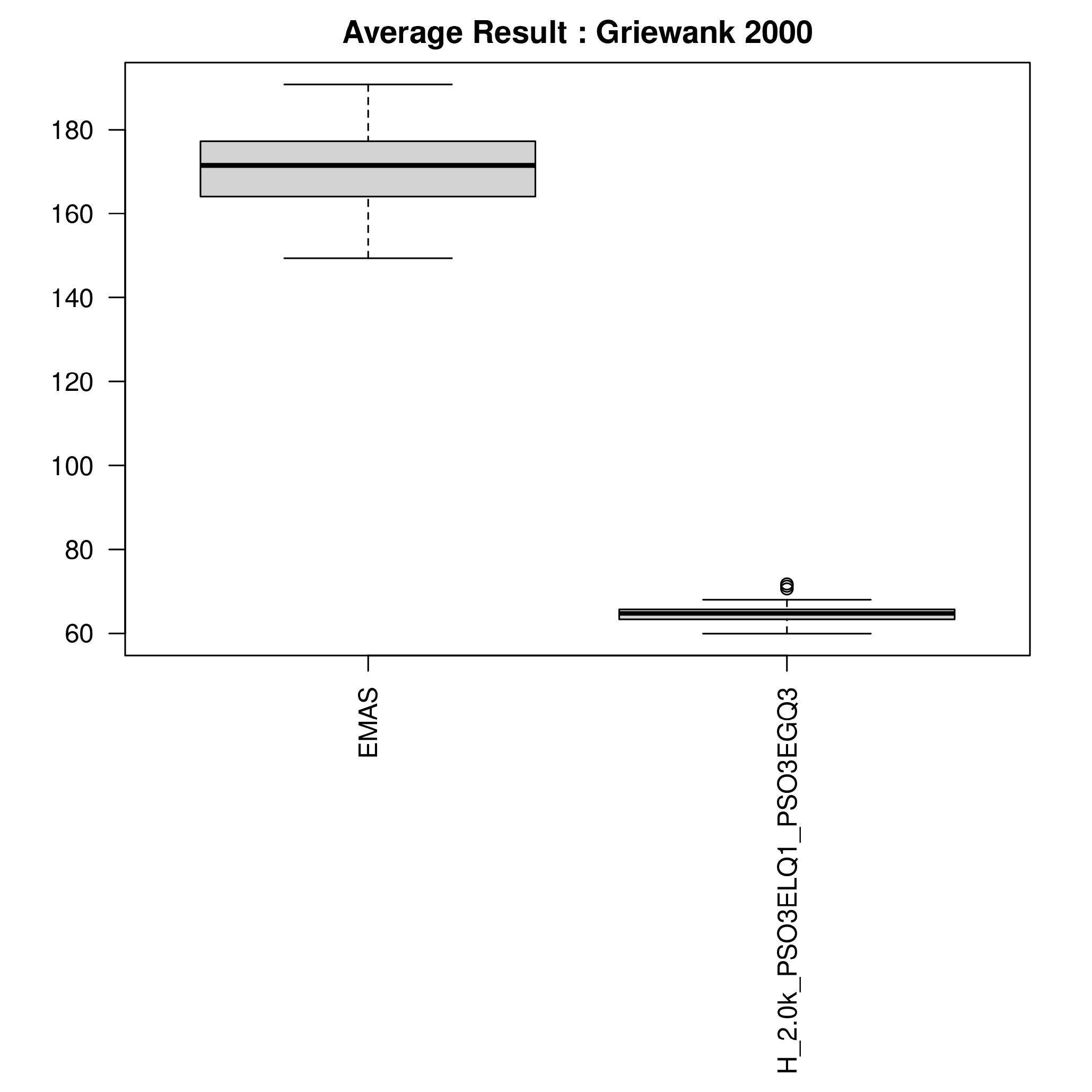}
            \caption{Griewank 2000D}
            \label{fig:z2_griewank_2000}
        \end{subfigure}
         \begin{subfigure}[b]{0.32\textwidth}
            \includegraphics[width=\textwidth]{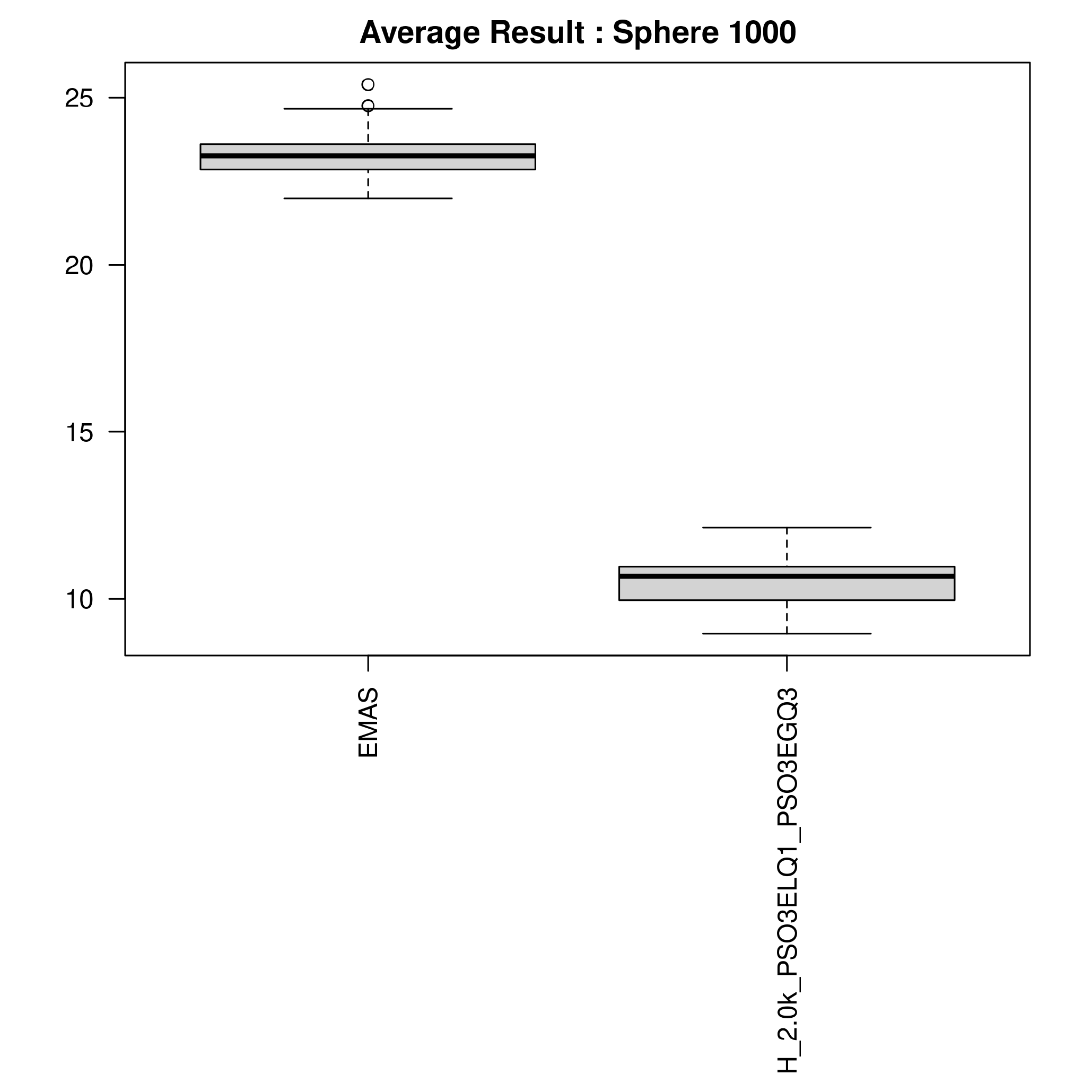}
            \caption{Sphere 1000D}
            \label{fig:z2_sphere_1000}
        \end{subfigure}
        \begin{subfigure}[b]{0.32\textwidth}
            \includegraphics[width=\textwidth]{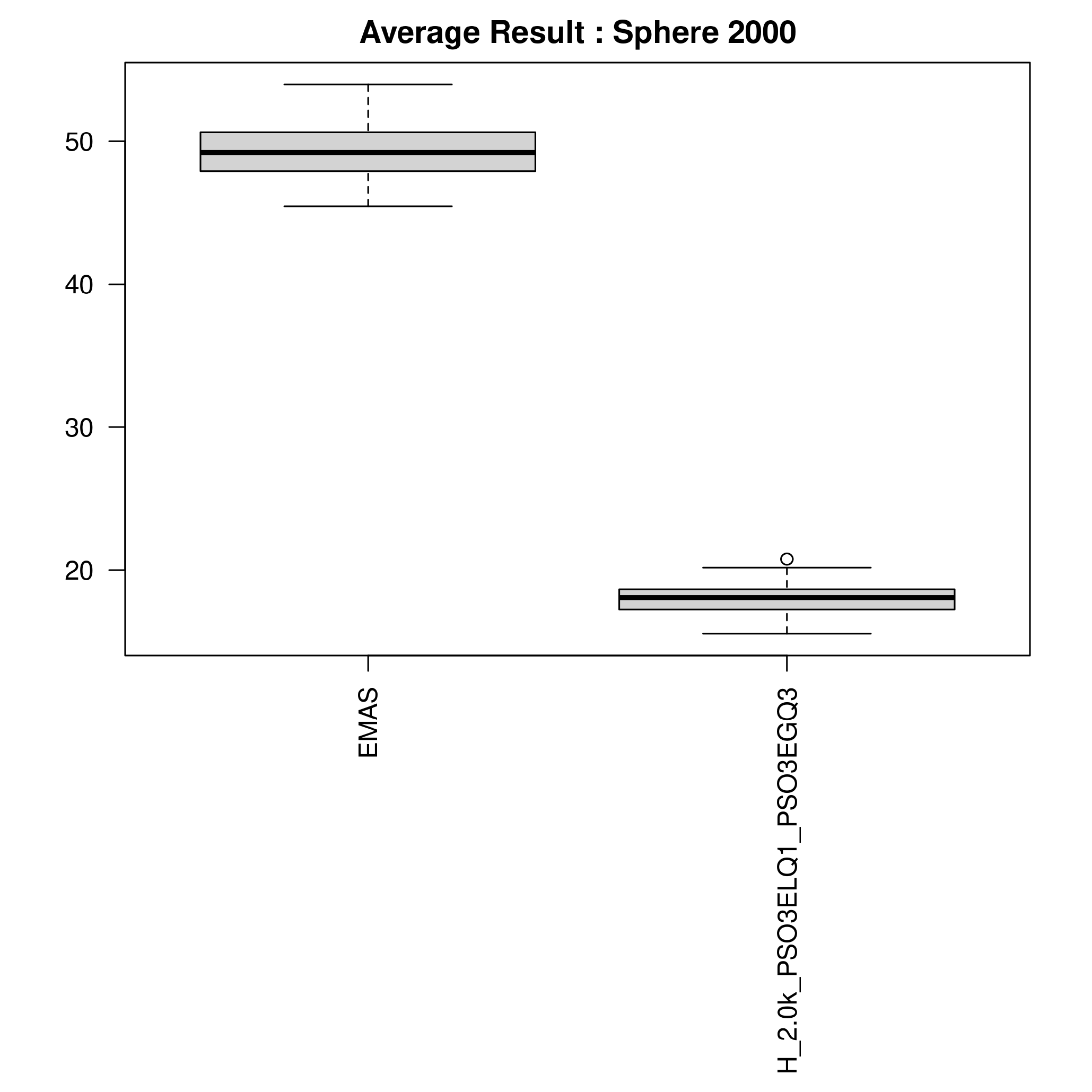}
            \caption{Sphere 2000D}
            \label{fig:z2_sphere_2000}
        \end{subfigure}
    \caption{Final fitnesses for selected benchmarks obtained by EMAS and HEMAS with two PSO hybridization operator (Rules ELQ1 and EGQ3)}
    \label{fig:final_z2_1}
\end{figure}

Taking a look at the graphs in Fig. \ref{fig:final_z2_1}, the dispersion of the final fitnesses for the observed cases was lower than it was previously.
The final fitnesses were visibly better in the observed case; e.g., solving the Rastrigin benchmark in 2000 dimensions reached ca. 2500, while this was approximately 4500 earlier (see Fig. \ref{fig:z2_rastrigin_2000}).

This observation can be confirmed when analyzing Table \ref{tab:Results2}. Generally, the results were similar; however, adding another hybridization algorithm allowed us to obtain significantly better results than in the case with a single operator (and in the case of the original EMAS). In only four cases (Griewank 100, Rastrigin 100, Rastrigin 300, and Sphere 100), HEMAS with two optimization operators achieved slightly worse results than HEMAS with one operator. In the remaining cases, adding a second operator helped improve the mean results, medians, and standard deviations as well as the minimum and maximum results.

\begingroup
\renewcommand{\arraystretch}{0.75}
\begin{table}[ht]
\centering
    \caption{Results of HEMAS with two optimizer operators for tested problems (values that were lower than those of HEMAS with one optimizer operator shown as \textbf{bold})}
    \begin{tabular}{llllll}
        \toprule
        \multicolumn{6}{c}{HEMAS with two optimizer operators } \\
            & {Mean} & {Median} & {SD} & {Minimum} & {Maximum} \\
        \midrule
        \textbf{Ackley   100}   & \textbf{4.965019} & \textbf{4.918697} & \textbf{0.23971}  & 4.438473 & \textbf{5.461094} \\
        \textbf{Ackley 300}     & \textbf{4.30319}  & \textbf{4.299941} & \textbf{0.16634}  & \textbf{3.943598} & \textbf{4.590263} \\
        \textbf{Ackley 500}     & \textbf{4.068269} & \textbf{4.048055} & \textbf{0.112506} & \textbf{3.854541} & \textbf{4.322338} \\
        \textbf{Ackley 1000}    & \textbf{3.933045} & \textbf{3.951304} & \textbf{0.103594} & \textbf{3.679261} & \textbf{4.097996} \\
        \textbf{Ackley 2000}    & \textbf{3.77687}  & \textbf{3.771572} & \textbf{0.049098} & \textbf{3.709228} & \textbf{3.922452} \\
        \textbf{Griewank 100}   & 6.546396 & 6.308251 & 1.00699  & 4.7822   & 8.400925 \\
        \textbf{Griewank 300}   & \textbf{14.16135} & \textbf{14.06403} & \textbf{1.19114}  & \textbf{11.98841} & \textbf{16.62449} \\
        \textbf{Griewank 500}   & \textbf{21.24983} & \textbf{20.98756} & \textbf{1.582526} & \textbf{18.60745} & \textbf{24.63775} \\
        \textbf{Griewank 1000}  & \textbf{37.00521} & \textbf{36.54933} & \textbf{2.064537} & \textbf{33.80159} & \textbf{40.41014} \\
        \textbf{Griewank 2000}  & \textbf{64.95195} & \textbf{64.75208} & \textbf{2.75928}  & \textbf{59.95819} & \textbf{71.76175} \\
        \textbf{Rastrigin 100}  & 218.1213 & 214.0323 & 30.01992 & 166.5907 & 285.4015 \\
        \textbf{Rastrigin 300}  & 739.861  & 736.5297 & \textbf{43.76646} & 638.4984 & 844.7063 \\
        \textbf{Rastrigin 500}  & \textbf{1001.599} & \textbf{1024.188} & 277.8012 & \textbf{619.1786} & 1401.642 \\
        \textbf{Rastrigin 1000} & \textbf{1289.217} & \textbf{1279.006} & \textbf{64.37773} & \textbf{1165.197} & \textbf{1457.02}  \\
        \textbf{Rastrigin 2000} & \textbf{2523.298} & \textbf{2513.266} & \textbf{94.30556} & \textbf{2356.164} & \textbf{2746.758} \\
        \textbf{Sphere 100}     & 1.634651 & 1.657542 & 0.274262 & 1.201514 & 2.092893 \\
        \textbf{Sphere 300}     & \textbf{4.047409} & \textbf{4.063365} & \textbf{0.467729} & \textbf{3.27734}  & \textbf{5.462412} \\
        \textbf{Sphere 500}     & \textbf{5.799361} & \textbf{5.696378} & \textbf{0.442802} & \textbf{4.990668} & \textbf{6.609995} \\
        \textbf{Sphere 1000}    & \textbf{10.5391}  & \textbf{10.68029} & \textbf{0.791381} & \textbf{8.964509} & \textbf{12.1381}  \\
        \textbf{Sphere 2000}    & \textbf{18.10579} & \textbf{18.08049} & \textbf{1.068876} & \textbf{15.57007} & \textbf{20.7757}  \\
        \bottomrule
    \end{tabular}
	\label{tab:Results2}
\end{table}
\endgroup

\subsection{HEMAS with three hybridization operators}
\begin{figure}[ht]
    \centering
    \includegraphics[width=0.98\textwidth]{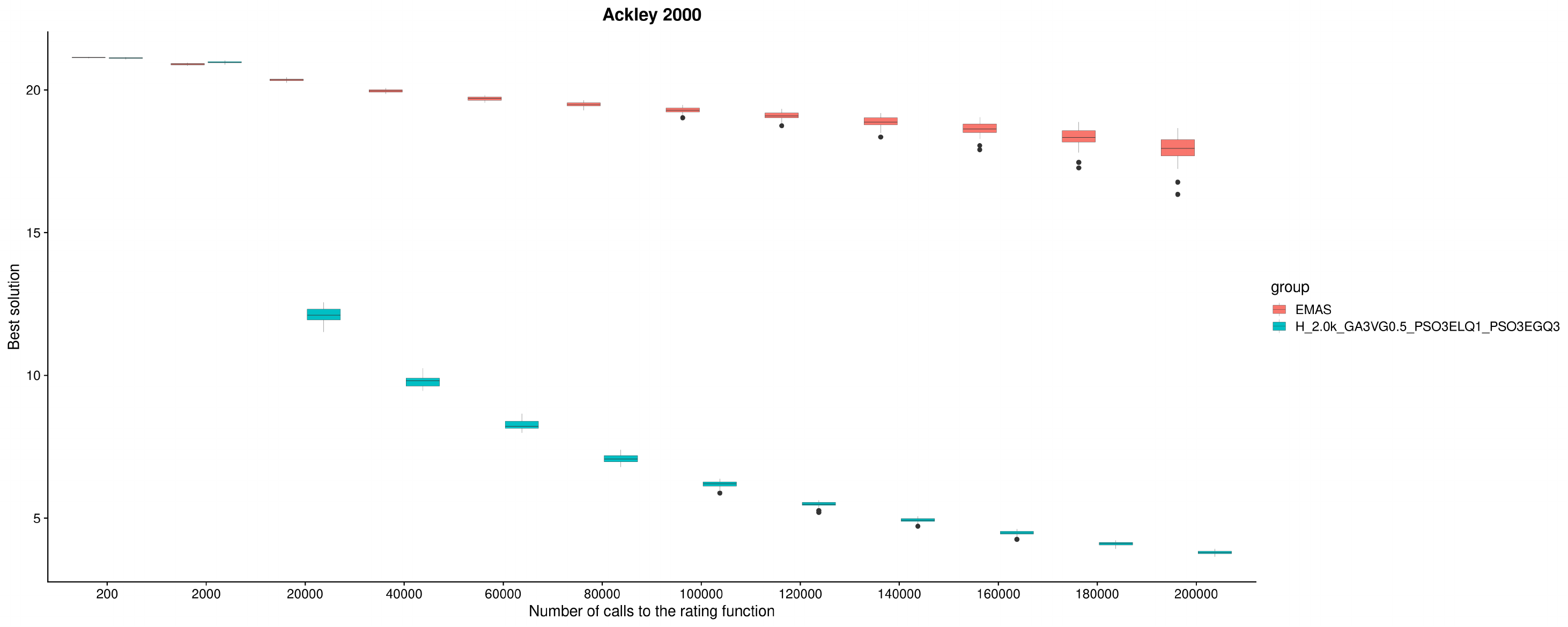}
        \caption{Results for EMAS (red [left]) and HEMAS with GA operator (Condition VG.0.5) and PSO (Conditions ELQ1 and EGQ3 -- blue [left]) for Ackley 2000 dimensions}
        \label{fig:partial_results_3}
\end{figure}

In Fig. \ref{fig:partial_results_3}, the results that were obtained for HEMAS with three hybridization operators are shown. HEMAS executed PSO as its hybrid step for those agents with energy levels that were lower than the first energy quartile as well as for those agents with energy levels that were higher than the third quartile. Moreover, the genetic algorithm was run for the whole population when then diversity was greater than $0.5$. Again, this combination proved to be the best of all of the tested HEMAS versions with three operators (ES + DE + SA, ES + PSO + SA, ES + ES + SA, ES + GA + SA, ES + NES + SA, ES + SA + SA, DE + PSO + PSO, ES + PSO + PSO, GA + PSO + PSO, NES + PSO + PSO, SA + PSO + PSO, and PSO + PSO + PSO) triggered by the conditions (VE0 + EL3 + EG17, VE0 + ELQ1 + EGQ3, VE0 + SLQ1 + SGQ3, VG0.5 + EL3 + EG17, VG0.5 + ELQ1 + EGQ3, VG0.5 + SLQ1 + SGQ3, VG1 + EL3 + EG17, VG1 + ELQ1 + EGQ3, and VG1 + SLQ1 + SGQ3).
The obtained results (Fig. \ref{fig:partial_results_3}) were quite similar to the previous ones (the experiment with two hybridization operators).

%Na wykresie \ref{fig:partial_results_3} poza umieszczonymi rozwiązaniami algorytmu EMAS (niezmiennie w kolorze czerwonym, po lewej stronie) umieszczono rozwiązania algorytmu HEMAS z trzema operatorami (po prawej, niebieski kolor). HEMAS w tym przypadku uruchamia operator PSO dla agentów z energią poniżej 1 kwartyla oraz dla agentów z energią powyżej 3 kwartyla energii w populacji a operator GA gdy różnorodność rozwiązań jest powyżej $0.5$.
%Ponownie na fig:\ref{fig:final_z3_1}, fig:\ref{fig:final_z3_2} oraz fig:\ref{fig:final_z3_3} można zobaczyć uśrednione wyniki końcowe algorytmu bazowego i tej wersji hybrydy. Ponownie, wszytkie wielkości problemu Ackley hybryda rozwiązuje lepiej. Przy problemie Rastrigin w wielkości 100D i 300D, Griewank 100D i Sphere 100D widać niewielkie pogorszenie wyników. Dla pozostałych problemów i rozmiarów hybryda działa lepiej.
%Ponownie na górze tabeli \ref{tab:Results3} można odczytać dokładne wartości uzyskane przez EMAS i porównać je z wartościami uzyskanymi przez HEMAS z 3 operatorami na dole.

\begin{figure}[p]
    \centering
        % \begin{subfigure}[b]{0.32\textwidth}
        %     \includegraphics[width=\textwidth]{graphics/z3/AverageAckley 100.pdf}
        %     \caption{100D funkcja Ackley}
        %     \label{fig:z3_ackley_100}
        % \end{subfigure}
        % \begin{subfigure}[b]{0.32\textwidth}
        %     \includegraphics[width=\textwidth]{graphics/z3/AverageAckley 300.pdf}
        %     \caption{300D funkcja Ackley}
        %     \label{fig:z3_ackley_300}
        % \end{subfigure}
        % \begin{subfigure}[b]{0.32\textwidth}
        %     \includegraphics[width=\textwidth]{graphics/z3/AverageAckley 500.pdf}
        %     \caption{500D funkcja Ackley}
        %     \label{fig:z3_ackley_500}
        % \end{subfigure}
        \begin{subfigure}[b]{0.32\textwidth}
            \includegraphics[width=\textwidth]{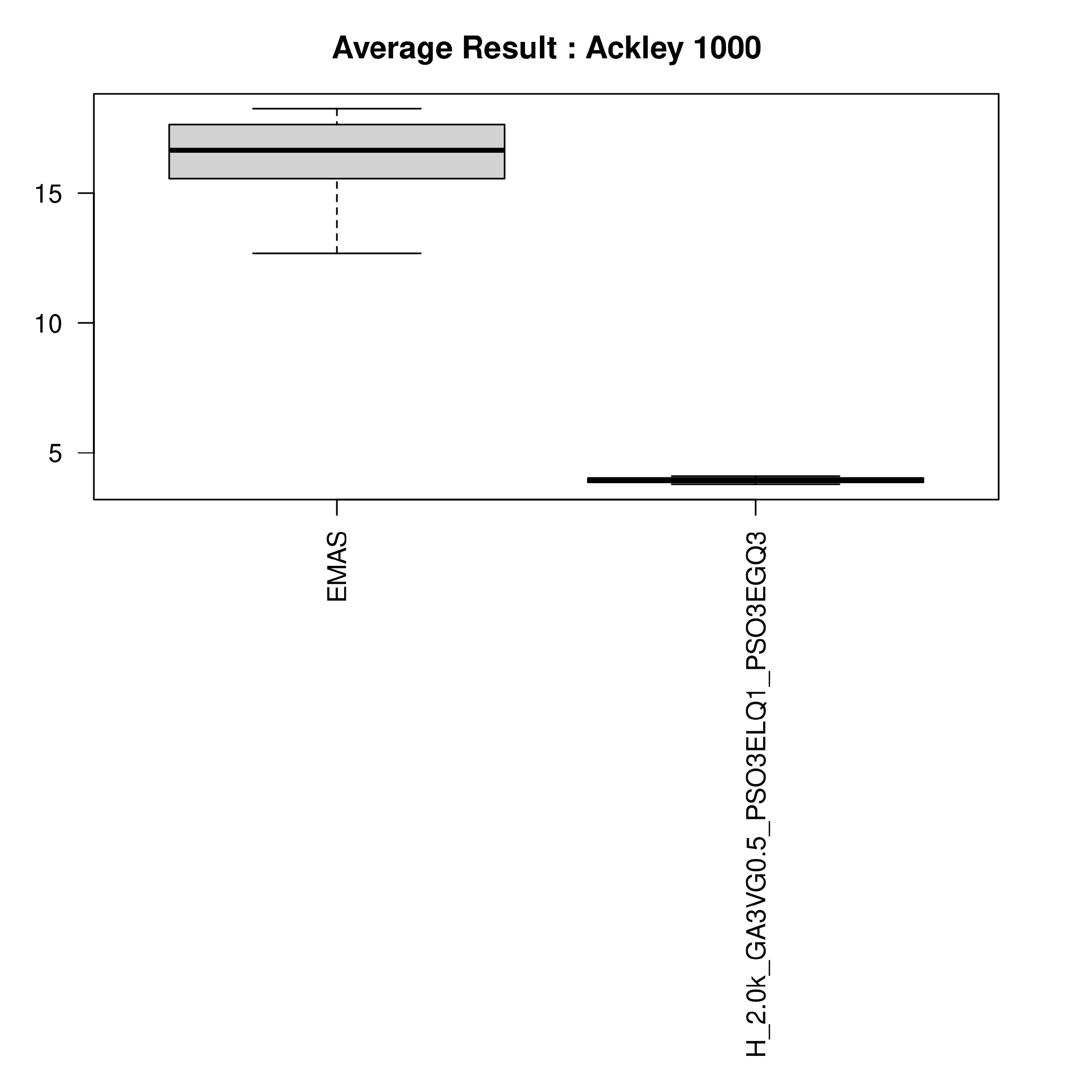}
            \caption{1000D funkcja Ackley}
            \label{fig:z3_ackley_1000}
        \end{subfigure}
        \begin{subfigure}[b]{0.32\textwidth}
            \includegraphics[width=\textwidth]{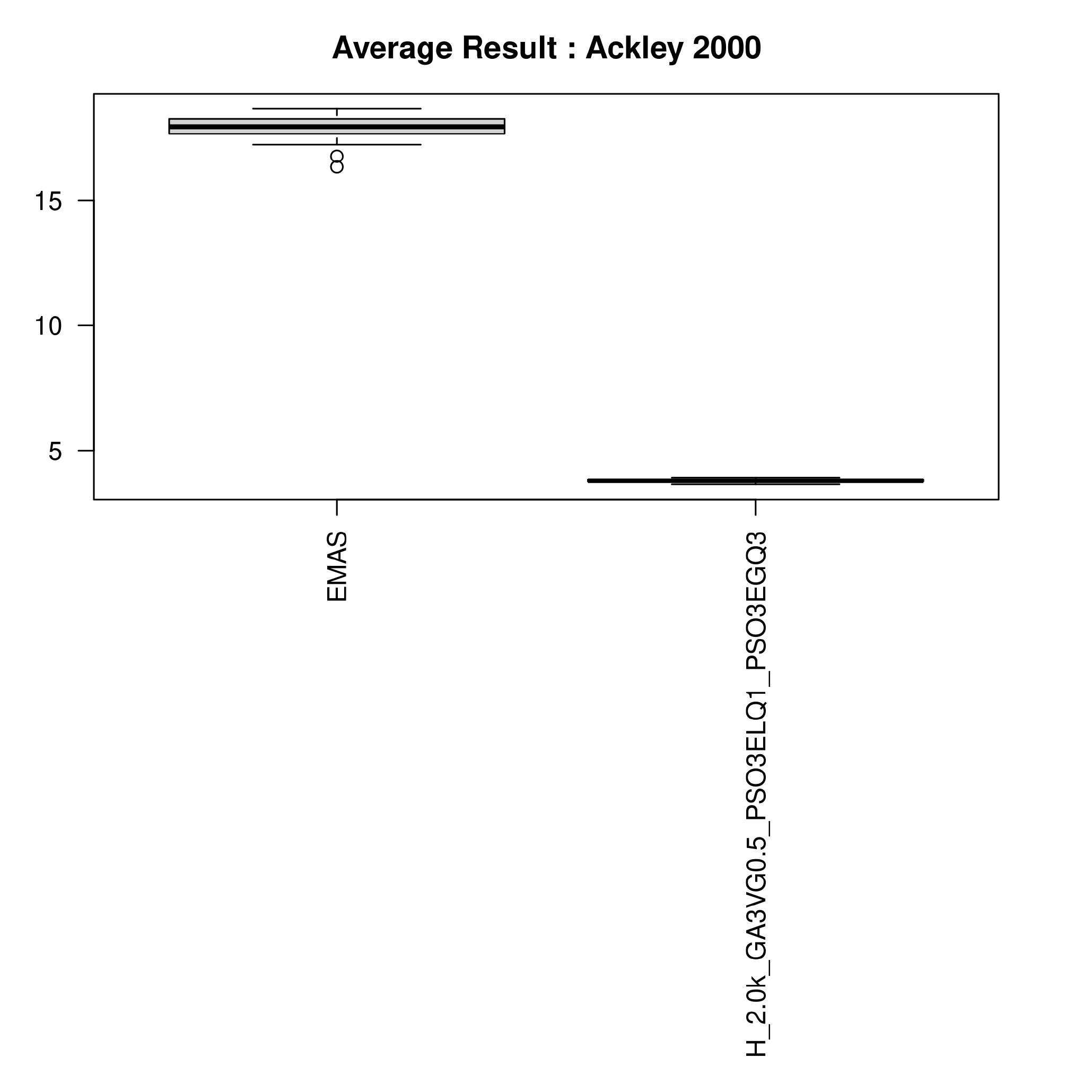}
            \caption{2000D funkcja Ackley}
            \label{fig:z3_ackley_2000}
        \end{subfigure}
        % \begin{subfigure}[b]{0.32\textwidth}
        %     \includegraphics[width=\textwidth]{graphics/z3/AverageRastrigin 100.pdf}
        %     \caption{100D funkcja Rastrigin}
        %     \label{fig:z3_rastrigin_100}
        % \end{subfigure}
        % \begin{subfigure}[b]{0.32\textwidth}
        %     \includegraphics[width=\textwidth]{graphics/z3/AverageRastrigin 300.pdf}
        %     \caption{300D funkcja Rastrigin}
        %     \label{fig:z3_rastrigin_300}
        % \end{subfigure}
        % \begin{subfigure}[b]{0.32\textwidth}
        %     \includegraphics[width=\textwidth]{graphics/z3/AverageRastrigin 500.pdf}
        %     \caption{500D funkcja Rastrigin}
        %     \label{fig:z3_rastrigin_500}
        % \end{subfigure}
        \begin{subfigure}[b]{0.32\textwidth}
            \includegraphics[width=\textwidth]{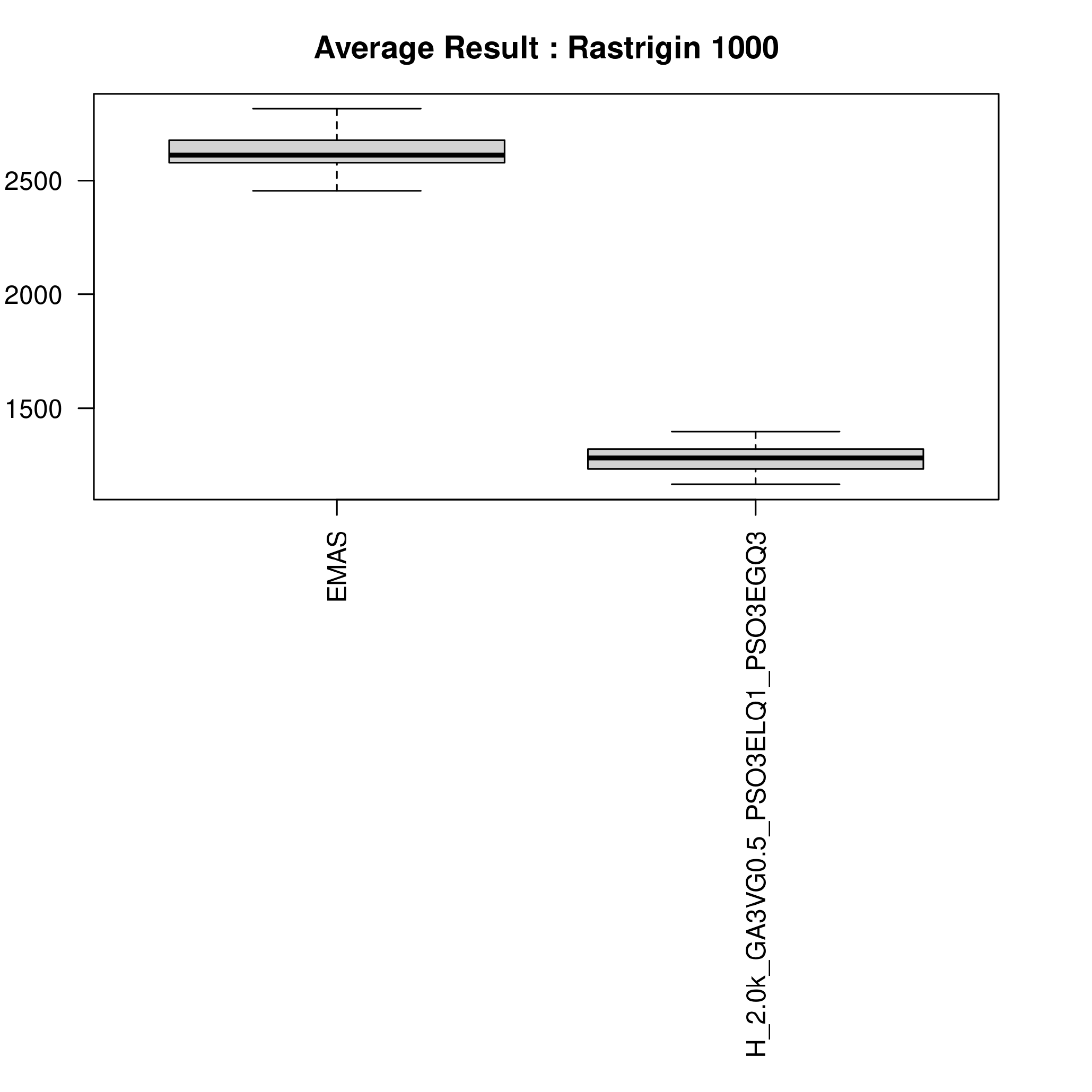}
            \caption{1000D funkcja Rastrigin}
            \label{fig:z3_rastrigin_1000}
        \end{subfigure}
         \begin{subfigure}[b]{0.32\textwidth}
            \includegraphics[width=\textwidth]{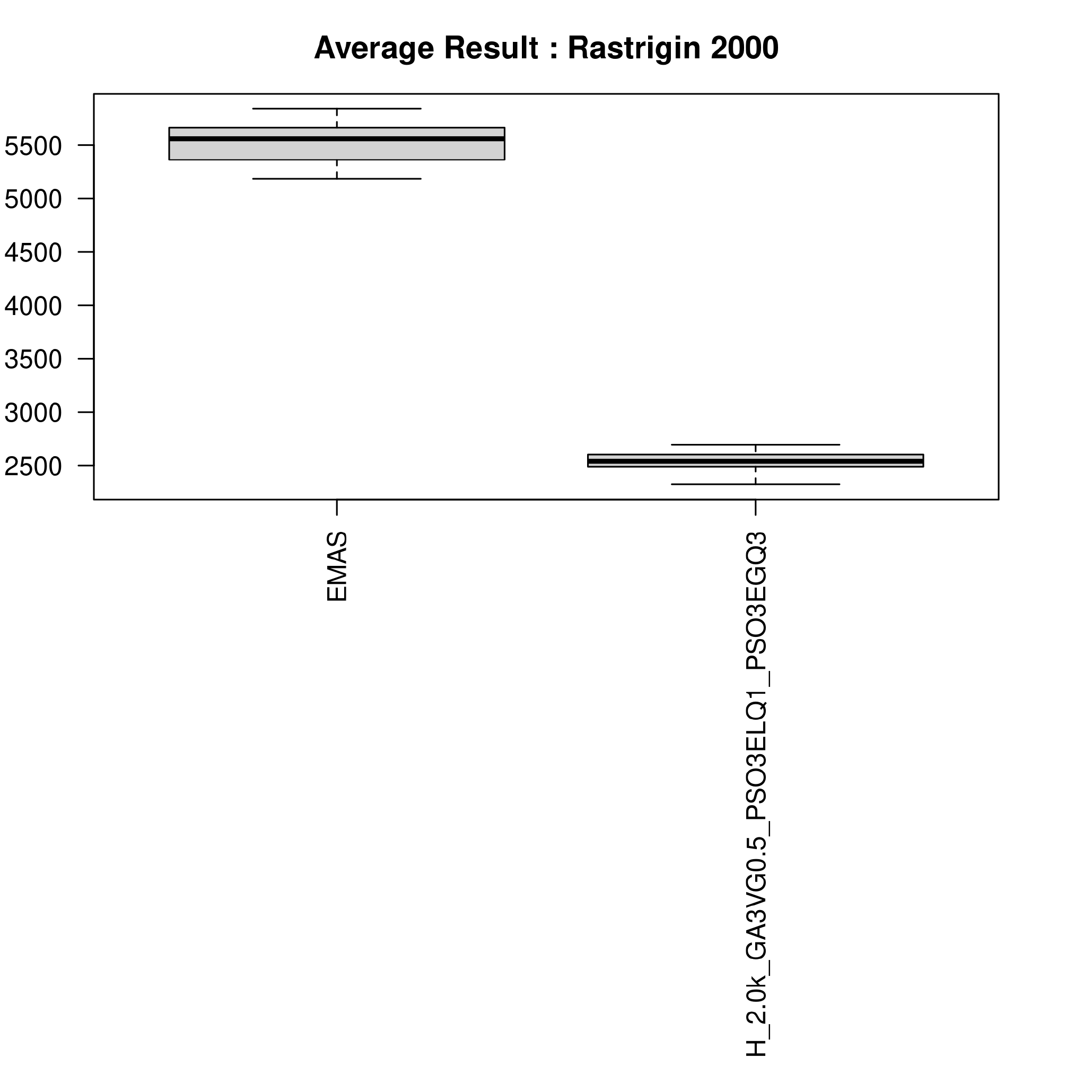}
            \caption{2000D funkcja Rastrigin}
            \label{fig:z3_rastrigin_2000}
        \end{subfigure}
        \begin{subfigure}[b]{0.32\textwidth}
            \includegraphics[width=\textwidth]{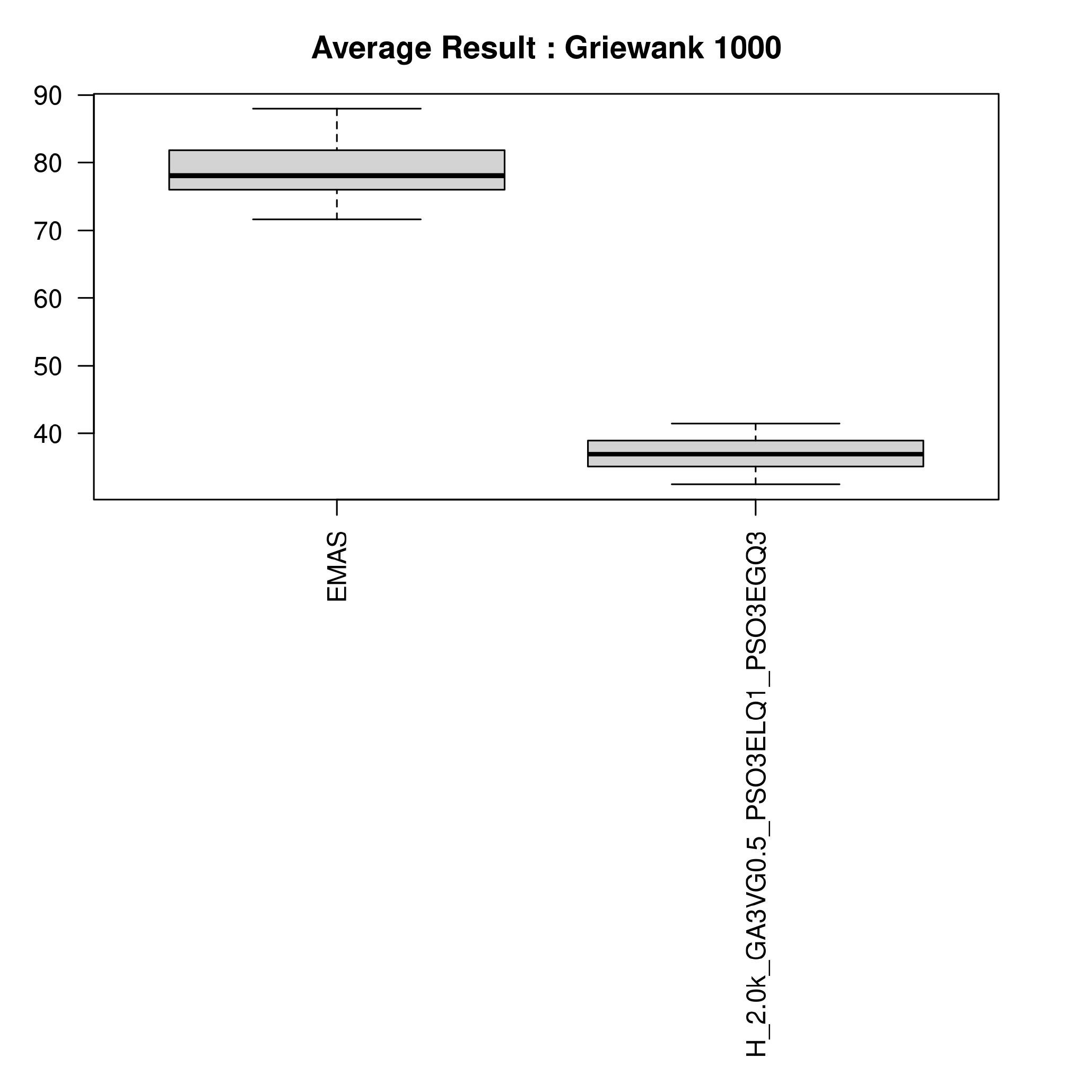}
            \caption{1000D funkcja Griewank}
            \label{fig:z3_griewank_1000}
        \end{subfigure}
        \begin{subfigure}[b]{0.32\textwidth}
            \includegraphics[width=\textwidth]{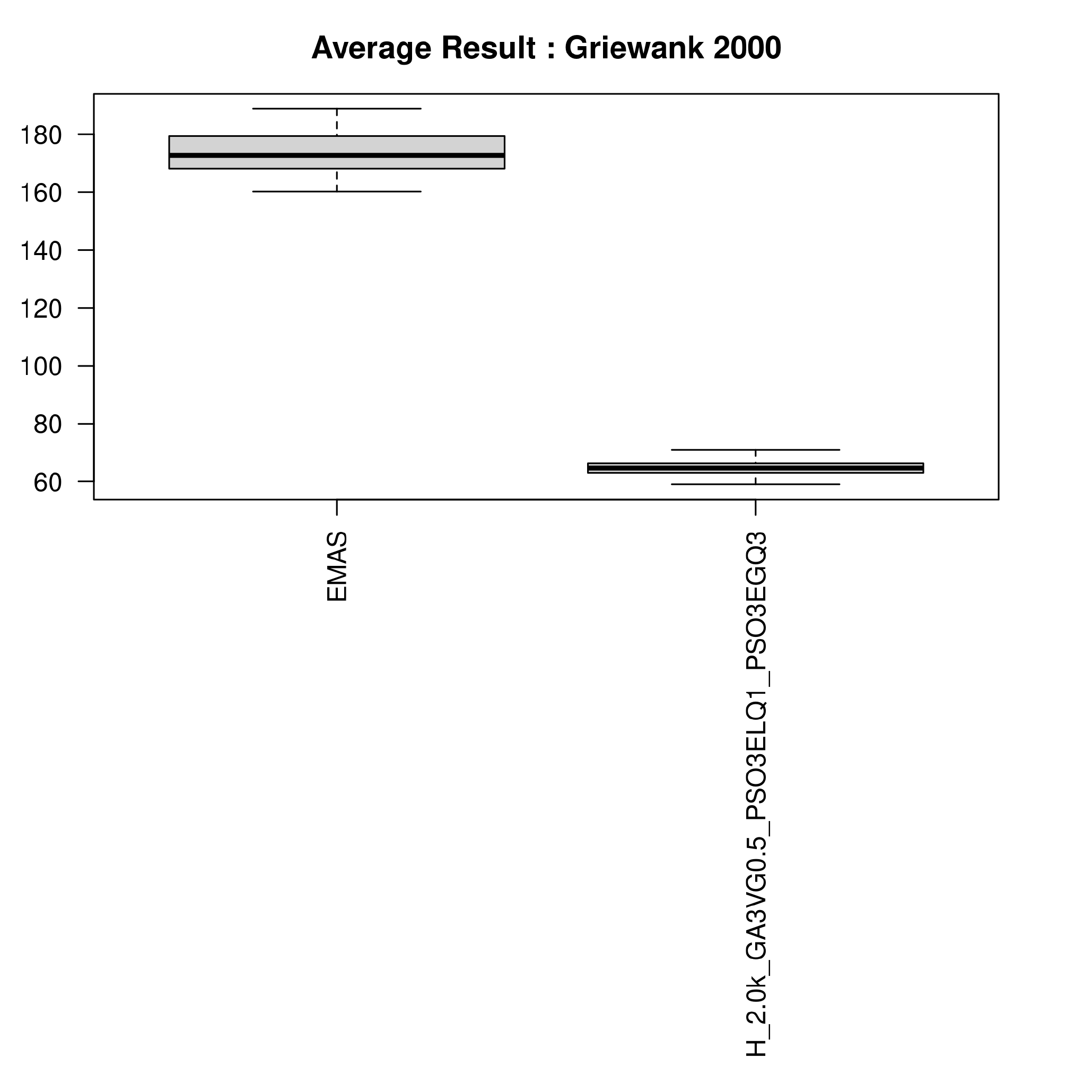}
            \caption{2000D funkcja Griewank}
            \label{fig:z3_griewank_2000}
        \end{subfigure}
         \begin{subfigure}[b]{0.32\textwidth}
            \includegraphics[width=\textwidth]{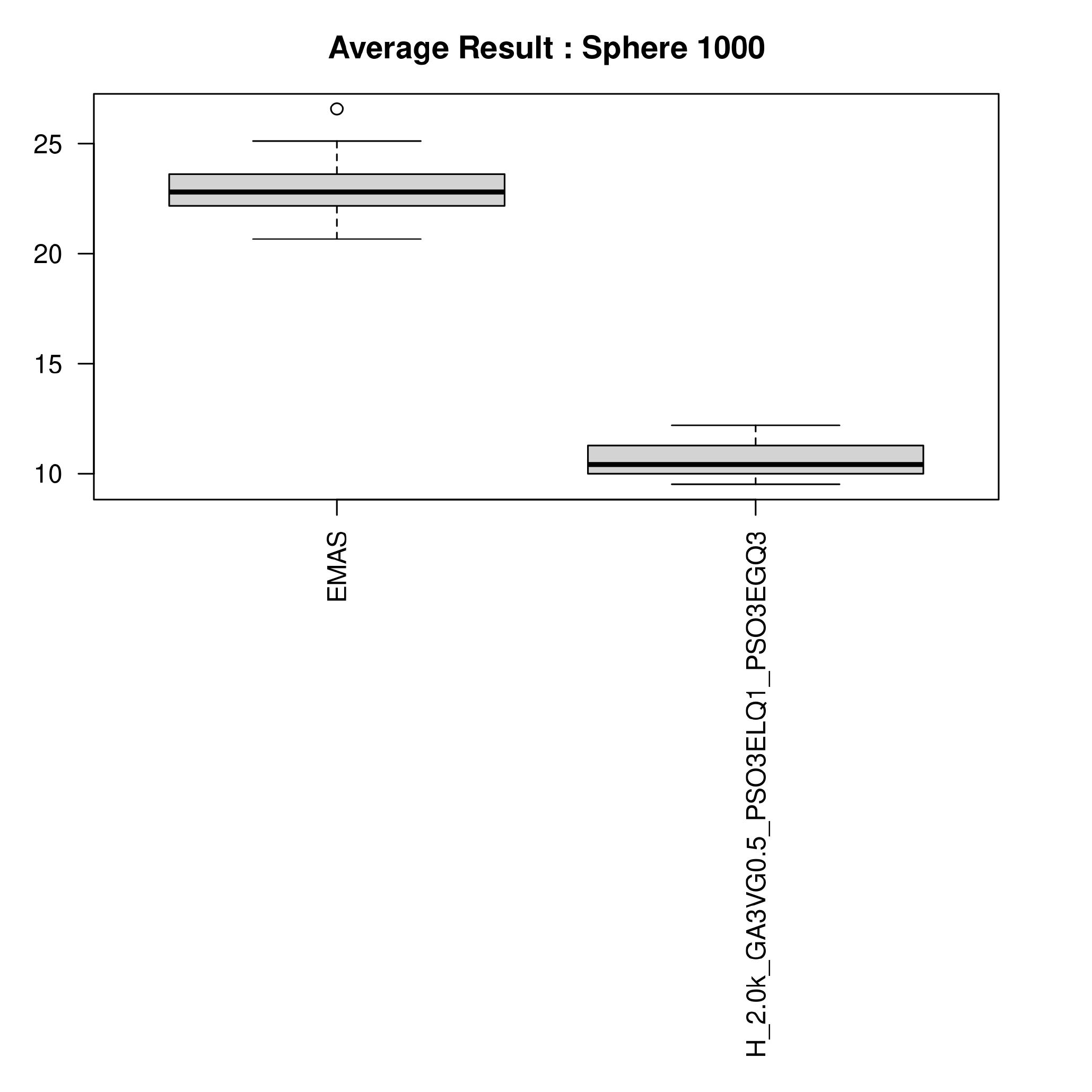}
            \caption{1000D funkcja Sphere}
            \label{fig:z3_sphere_1000}
        \end{subfigure}
        \begin{subfigure}[b]{0.32\textwidth}
            \includegraphics[width=\textwidth]{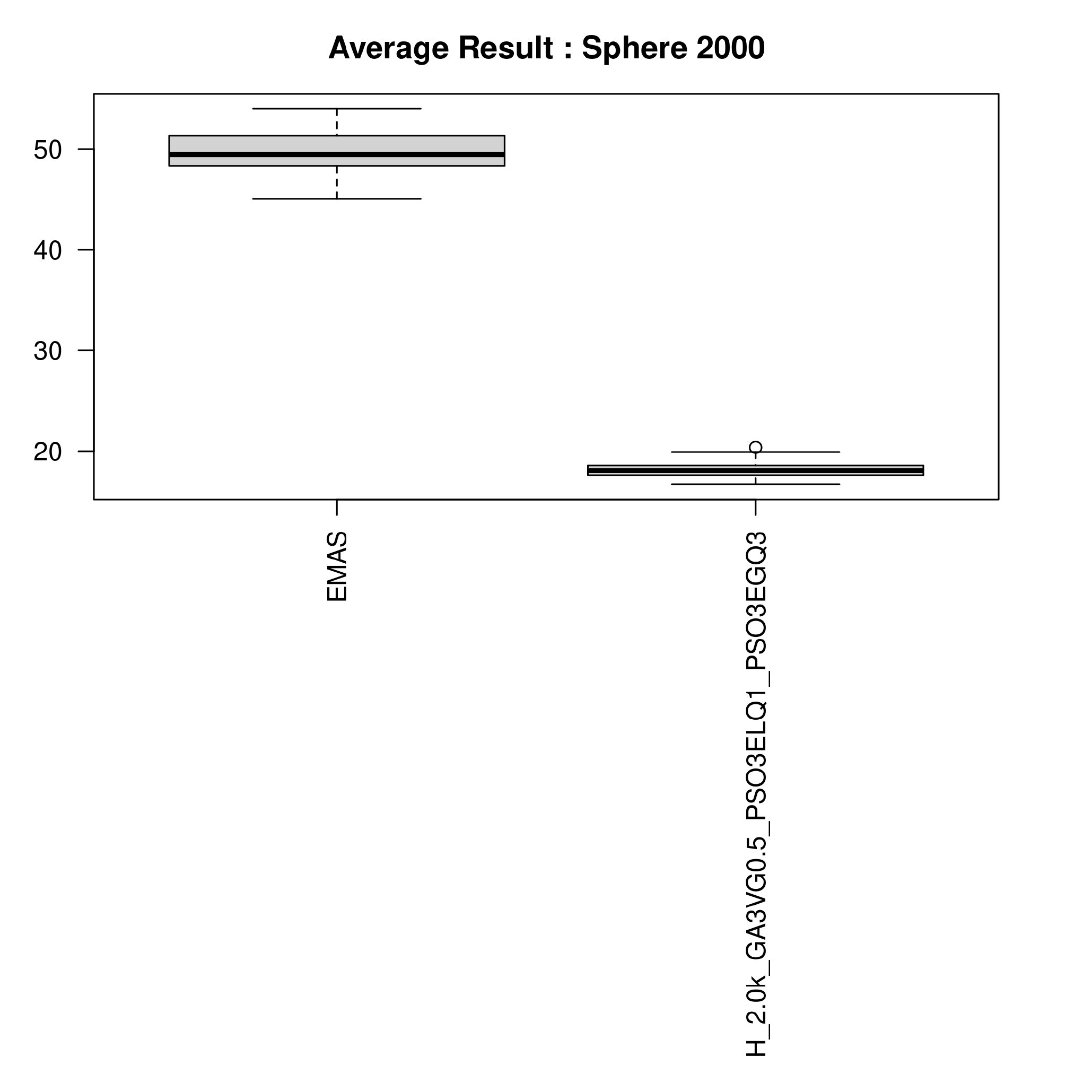}
            \caption{2000D funkcja Sphere}
            \label{fig:z3_sphere_2000}
        \end{subfigure}
    \caption{Final fitnesses for selected benchmarks that were obtained by EMAS and HEMAS with two PSO hybridization operators (Rules ELQ1 and EGQ3) and GA operator (Rule VG0.5)}
	 \label{fig:final_z3_1}
\end{figure}

The final fitnesses that are displayed using box-and-whisker plots (Fig. \ref{fig:final_z3_1}) were similar to the results that were obtained in the previous experiment.

Finally, Table \ref{tab:Results3} presents the final fitnesses for all of the benchmarks that were tackled; again, it is easy to see that HEMAS beat EMAS (although these results were not statistically different to those that were presented for two hybridization operators. Comparing the values that were obtained by HEMAS with two and three operators, it can be seen that adding the next operator clearly helped in a few cases (Griewank 2000, Rastrigin 300, Rastrigin 500, and Sphere 300); in one case (Sphere 500), it worsened the results.
In the remaining test cases, single values (mean, median, standard deviation, minimum, or maximum) were better; however, this did not allow for unequivocally recognizing this version as being significantly better.

\begingroup
\renewcommand{\arraystretch}{0.75}
\begin{table}[ht]
\centering
    \caption{Results of HEMAS with three optimizer operators for tested problems (values that were lower than those of HEMAS with two optimizer operators shown as \textbf{bold})}
    \begin{tabular}{llllll}
        \toprule
        \multicolumn{6}{c}{HEMAS with three optimizer operators } \\
            & {Mean} & {Median} & {SD} & {Minimum} & {Maximum} \\
        \midrule
        \textbf{Ackley 100}     & \textbf{4.885853} & \textbf{4.899817} & 0.249164 & \textbf{4.198643} & 5.467292 \\
        \textbf{Ackley 300}     & 4.313266 & 4.308561 & \textbf{0.154479} & \textbf{3.912178} & \textbf{4.579967} \\
        \textbf{Ackley 500}     & 4.170135 & 4.16319  & \textbf{0.095219} & 4.007301 & 4.388715 \\
        \textbf{Ackley 1000}    & 3.94585  & \textbf{3.949122} & \textbf{0.097332} & 3.785541 & 4.111698 \\
        \textbf{Ackley 2000}    & 3.795452 & 3.789527 & 0.064746 & \textbf{3.642467} & \textbf{3.921606} \\
        \textbf{Griewank 100}   & 6.703274 & 6.551771 & \textbf{0.905072} & \textbf{4.730231} & 8.567404 \\
        \textbf{Griewank 300}   & 14.77065 & 14.83329 & 1.404183 & \textbf{11.65218} & 17.91965 \\
        \textbf{Griewank 500}   & 21.30266 & 21.03757 & 1.786992 & \textbf{17.40044} & \textbf{24.39079} \\
        \textbf{Griewank 1000}  & \textbf{36.92428} & 36.93708 & 2.323852 & \textbf{32.45933} & 41.48806 \\
        \textbf{Griewank 2000}  & \textbf{64.41718} & \textbf{64.67174} & 3.002067 & \textbf{59.00248} & \textbf{71.02364} \\
        \textbf{Rastrigin 100}  & 219.6632 & 215.7575 & \textbf{22.13633} & 176.4731 & \textbf{259.0824} \\
        \textbf{Rastrigin 300}  & \textbf{722.2067} & \textbf{719.6956} & 73.87167 & \textbf{412.456}  & \textbf{830.8502} \\
        \textbf{Rastrigin 500}  & \textbf{937.0896} & \textbf{776.0801} & \textbf{266.8731} & \textbf{603.3675} & \textbf{1362.971} \\
        \textbf{Rastrigin 1000} & \textbf{1280.509} & 1281.532 & \textbf{58.12613} & 1165.998 & \textbf{1397.216} \\
        \textbf{Rastrigin 2000} & 2534.718 & 2539.554 & \textbf{85.93249} & \textbf{2323.532} & \textbf{2696.213} \\
        \textbf{Sphere 100}     & 1.721324 & 1.702814 & 0.326905 & \textbf{1.152834} & 2.374062 \\
        \textbf{Sphere 300}     & \textbf{3.922142} & \textbf{3.871625} & \textbf{0.4563}   & \textbf{3.096756} & \textbf{4.74739}  \\
        \textbf{Sphere 500}     & 6.034614 & 5.922964 & 0.537151 & 5.026456 & 7.33136  \\
        \textbf{Sphere 1000}    & 10.57195 & \textbf{10.41114} & \textbf{0.726729} & 9.506711 & 12.20145 \\
        \textbf{Sphere 2000}    & 18.1306  & \textbf{18.07885} & \textbf{0.842056} & 16.69438 & \textbf{20.38125} \\
        \bottomrule
    \end{tabular}
	\label{tab:Results3}
\end{table}
\endgroup

After finishing the experiments, we checked the statistical significance of the findings. We assumed a significance level of $\alpha=0.05$ and applied the Kruskal-Wallis test, whose results indicated that our final fitnesses (computed for all of the experiments for all of the tested algorithms) came from different distributions. Then, we applied Dunn's test in order to compare them pairwise. The results of these comparisons can be seen in Tables \ref{tab:Dunn1} and
\ref{tab:Dunn2}. All of the \emph{p-values} that were lower than the significance level (pointing out that the null hypothesis has a very low probability) are typeset in bold. Apparently, the results that were produced by both versions of the HEMAS algorithm were statistically different than with EMAS; however, the observed means were not different when comparing both HEMAS versions. It can be interpreted that the tested three-hybridization-operator configuration was not completely necessary; in the tested case, the two-operator version was enough to beat EMAS and the one-operator version. More-complex methods do not always bring better results.

% DUNN table
\begin{table}[p]
\centering
\caption{Dunn's test p-values for Rastrigin and Ackley benchmarks for EMAS and all HEMAS versions (values that were lower than significance level shown as \textbf{bold})}
\resizebox{\textwidth}{!}{%
\begin{tabular}{|l|llll|}
    \hline
    \textbf{Rastrigin 100} & \multicolumn{1}{l|}{\textbf{EMAS}} & \multicolumn{1}{l|}{\textbf{HEMAS with one operator}} & \multicolumn{1}{l|}{\textbf{HEMAS with two operators}} & \textbf{HEMAS with three operators} \\ \hline
    \textbf{EMAS} &  & \textbf{1.976778e-02} & 7.303993e-02 & \textbf{1.807253e-02} \\ \cline{1-1}
    \textbf{HEMAS with one operator} &  &  & \textbf{3.734702e-05} & \textbf{2.667955e-06} \\ \cline{1-1}
    \textbf{HEMAS with two operators} &  &  &  & 5.676285e-01 \\ \hline
    \textbf{Rastrigin 300} & \multicolumn{1}{l|}{\textbf{EMAS}} & \multicolumn{1}{l|}{\textbf{HEMAS with one operator}} & \multicolumn{1}{l|}{\textbf{HEMAS with two operators}} & \textbf{HEMAS with three operators} \\ \hline
    \textbf{EMAS} &  & \textbf{8.599426e-07} & \textbf{2.546784e-02} & 2.069998e-01 \\ \cline{1-1}
    \textbf{HEMAS with one operator} &  &  & \textbf{8.338118e-13} & \textbf{6.285251e-10} \\ \cline{1-1}
    \textbf{HEMAS with two operators} &  &  &  & 3.308649e-01 \\ \hline
    \textbf{Rastrigin 500} & \multicolumn{1}{l|}{\textbf{EMAS}} & \multicolumn{1}{l|}{\textbf{HEMAS with one operator}} & \multicolumn{1}{l|}{\textbf{HEMAS with two operators}} & \textbf{HEMAS with three operators} \\ \hline
    \textbf{EMAS} &  & \textbf{4.455709e-05} & \textbf{2.218060e-04} & \textbf{2.667955e-06} \\ \cline{1-1}
    \textbf{HEMAS with one operator} &  &  & 6.967646e-01 & 5.402914e-01 \\ \cline{1-1}
    \textbf{HEMAS with two operators} &  &  &  & 3.163127e-01 \\ \hline
    \textbf{Rastrigin 1000} & \multicolumn{1}{l|}{\textbf{EMAS}} & \multicolumn{1}{l|}{\textbf{HEMAS with one operator}} & \multicolumn{1}{l|}{\textbf{HEMAS with two operators}} & \textbf{HEMAS with three operators} \\ \hline
    \textbf{EMAS} &  & \textbf{5.573688e-04} & \textbf{1.962565e-16} & \textbf{5.992220e-17} \\ \cline{1-1}
    \textbf{HEMAS with one operator} &  &  & \textbf{1.816879e-06} & \textbf{8.931654e-07} \\ \cline{1-1}
    \textbf{HEMAS with two operators} &  &  &  & 8.878453e-01 \\ \hline
    \textbf{Rastrigin 2000} & \multicolumn{1}{l|}{\textbf{EMAS}} & \multicolumn{1}{l|}{\textbf{HEMAS with one operator}} & \multicolumn{1}{l|}{\textbf{HEMAS with two operators}} & \textbf{HEMAS with three operators} \\ \hline
    \textbf{EMAS} &  & \textbf{2.880485e-03} & \textbf{5.992220e-17} & \textbf{1.814335e-15} \\ \cline{1-1}
    \textbf{HEMAS with one operator} &  &  & \textbf{7.237778e-08} & \textbf{6.585446e-07} \\ \cline{1-1}
    \textbf{HEMAS with two operators} &  &  &  & 6.803690e-01 \\ \hline
    \textbf{Ackley 100} & \multicolumn{1}{l|}{\textbf{EMAS}} & \multicolumn{1}{l|}{\textbf{HEMAS with one operator}} & \multicolumn{1}{l|}{\textbf{HEMAS with two operators}} & \textbf{HEMAS with three operators} \\ \hline
    \textbf{EMAS} &  & 2.043408e-01 & \textbf{4.822134e-06} & \textbf{6.262286e-08} \\ \cline{1-1}
    \textbf{HEMAS with one operator} &  &  & \textbf{9.562235e-04} & \textbf{3.444939e-05} \\ \cline{1-1}
    \textbf{HEMAS with two operators} &  &  &  & 4.016014e-01 \\ \hline
    \textbf{Ackley 300} & \multicolumn{1}{l|}{\textbf{EMAS}} & \multicolumn{1}{l|}{\textbf{HEMAS with one operator}} & \multicolumn{1}{l|}{\textbf{HEMAS with two operators}} & \textbf{HEMAS with three operators} \\ \hline
    \textbf{EMAS} &  & \textbf{3.733642e-02} & \textbf{4.266755e-14} & \textbf{8.189907e-14} \\ \cline{1-1}
    \textbf{HEMAS with one operator} &  &  & \textbf{4.486991e-08} & \textbf{7.237778e-08} \\ \cline{1-1}
    \textbf{HEMAS with two operators} &  &  &  & 9.319744e-01 \\ \hline
    \textbf{Ackley 500} & \multicolumn{1}{l|}{\textbf{EMAS}} & \multicolumn{1}{l|}{\textbf{HEMAS with one operator}} & \multicolumn{1}{l|}{\textbf{HEMAS with two operators}} & \textbf{HEMAS with three operators} \\ \hline
    \textbf{EMAS} &  & \textbf{1.263033e-02} & \textbf{1.566974e-18} & \textbf{1.548019e-12} \\ \cline{1-1}
    \textbf{HEMAS with one operator} &  &  & \textbf{3.159663e-10} & \textbf{4.737418e-06} \\ \cline{1-1}
    \textbf{HEMAS with two operators} &  &  &  & 8.641073e-02 \\ \hline
    \textbf{Ackley 1000} & \multicolumn{1}{l|}{\textbf{EMAS}} & \multicolumn{1}{l|}{\textbf{HEMAS with one operator}} & \multicolumn{1}{l|}{\textbf{HEMAS with two operators}} & \textbf{HEMAS with three operators} \\ \hline
    \textbf{EMAS} &  & \textbf{7.724534e-04} & \textbf{3.851370e-17} & \textbf{1.629437e-16} \\ \cline{1-1}
    \textbf{HEMAS with one operator} &  &  & \textbf{4.307154e-07} & \textbf{1.038844e-06} \\ \cline{1-1}
    \textbf{HEMAS with two operators} &  &  &  & 8.644424e-01 \\ \hline
    \textbf{Ackley 2000} & \multicolumn{1}{l|}{\textbf{EMAS}} & \multicolumn{1}{l|}{\textbf{HEMAS with one operator}} & \multicolumn{1}{l|}{\textbf{HEMAS with two operators}} & \textbf{HEMAS with three operators} \\ \hline
    \textbf{EMAS} &  & \textbf{8.371407e-04} & \textbf{2.039766e-18} & \textbf{1.926362e-15} \\ \cline{1-1}
    \textbf{HEMAS with one operator} &  &  & \textbf{6.133775e-08} & \textbf{4.109168e-06} \\ \cline{1-1}
    \textbf{HEMAS with two operators} &  &  &  & 4.184726e-01 \\ \hline
\end{tabular}%
    }
	\label{tab:Dunn1}
\end{table}

 \begin{table}[p]
\centering
\caption{Dunn's test p-values for Sphere and Griewank benchmarks for EMAS and all HEMAS versions (values that were lower than significance level shown as \textbf{bold})}
\resizebox{\textwidth}{!}{%
\begin{tabular}{|l|llll|}
    \hline
    \textbf{Sphere 100} & \multicolumn{1}{l|}{\textbf{EMAS}} & \multicolumn{1}{l|}{\textbf{HEMAS with one operator}} & \multicolumn{1}{l|}{\textbf{HEMAS with two operators}} & \textbf{HEMAS with three operators} \\ \hline
    \textbf{EMAS} &  & \textbf{2.185899e-04} & 2.349787e-01 & 6.350003e-02 \\ \cline{1-1}
    \textbf{HEMAS with one operator} &  &  & \textbf{1.038844e-06} & \textbf{2.821341e-08} \\ \cline{1-1}
    \textbf{HEMAS with two operators} &  &  &  & 5.041064e-01 \\ \hline
    \textbf{Sphere 300} & \multicolumn{1}{l|}{\textbf{EMAS}} & \multicolumn{1}{l|}{\textbf{HEMAS with one operator}} & \multicolumn{1}{l|}{\textbf{HEMAS with two operators}} & \textbf{HEMAS with three operators} \\ \hline
    \textbf{EMAS} &  & \textbf{9.448358e-05} & \textbf{1.428577e-12} & \textbf{9.240425e-15} \\ \cline{1-1}
    \textbf{HEMAS with one operator} &  &  & \textbf{1.488512e-03} & \textbf{1.205738e-04} \\ \cline{1-1}
    \textbf{HEMAS with two operators} &  &  &  & 5.041064e-01 \\ \hline
    \textbf{Sphere 500} & \multicolumn{1}{l|}{\textbf{EMAS}} & \multicolumn{1}{l|}{\textbf{HEMAS with one operator}} & \multicolumn{1}{l|}{\textbf{HEMAS with two operators}} & \textbf{HEMAS with three operators} \\ \hline
    \textbf{EMAS} &  & \textbf{1.205738e-04} & \textbf{1.794020e-17} & \textbf{4.647410e-14} \\ \cline{1-1}
    \textbf{HEMAS with one operator} &  &  & \textbf{3.139845e-06} & \textbf{2.185899e-04} \\ \cline{1-1}
    \textbf{HEMAS with two operators} &  &  &  & 3.345696e-01 \\ \hline
    \textbf{Sphere 1000} & \multicolumn{1}{l|}{\textbf{EMAS}} & \multicolumn{1}{l|}{\textbf{HEMAS with one operator}} & \multicolumn{1}{l|}{\textbf{HEMAS with two operators}} & \textbf{HEMAS with three operators} \\ \hline
    \textbf{EMAS} &  & \textbf{5.131016e-04} & \textbf{1.579619e-16} & \textbf{1.021273e-16} \\ \cline{1-1}
    \textbf{HEMAS with one operator} &  &  & \textbf{1.783681e-06} & \textbf{1.375903e-06} \\ \cline{1-1}
    \textbf{HEMAS with two operators} &  &  &  & 9.585615e-01 \\ \hline
    \textbf{Sphere 2000} & \multicolumn{1}{l|}{\textbf{EMAS}} & \multicolumn{1}{l|}{\textbf{HEMAS with one operator}} & \multicolumn{1}{l|}{\textbf{HEMAS with two operators}} & \textbf{HEMAS with three operators} \\ \hline
    \textbf{EMAS} &  & \textbf{7.028607e-04} & \textbf{1.021273e-16} & \textbf{7.704896e-17} \\ \cline{1-1}
    \textbf{HEMAS with one operator} &  &  & \textbf{8.931654e-07} & \textbf{7.527817e-07} \\ \cline{1-1}
    \textbf{HEMAS with two operators} &  &  &  & 9.733539e-01 \\ \hline
    \textbf{Griewank 100} & \multicolumn{1}{l|}{\textbf{EMAS}} & \multicolumn{1}{l|}{\textbf{HEMAS with one operator}} & \multicolumn{1}{l|}{\textbf{HEMAS with two operators}} & \textbf{HEMAS with three operators} \\ \hline
    \textbf{EMAS} &  & \textbf{5.823958e-03} & 2.123927e-01 & \textbf{4.234562e-02} \\ \cline{1-1}
    \textbf{HEMAS with one operator} &  &  & \textbf{6.213707e-05} & \textbf{1.687548e-06} \\ \cline{1-1}
    \textbf{HEMAS with two operators} &  &  &  & 4.335717e-01 \\ \hline
    \textbf{Griewank 300} & \multicolumn{1}{l|}{\textbf{EMAS}} & \multicolumn{1}{l|}{\textbf{HEMAS with one operator}} & \multicolumn{1}{l|}{\textbf{HEMAS with two operators}} & \textbf{HEMAS with three operators} \\ \hline
    \textbf{EMAS} &  & \textbf{1.552909e-02} & \textbf{4.237878e-16} & \textbf{1.722625e-12} \\ \cline{1-1}
    \textbf{HEMAS with one operator} &  &  & \textbf{1.118106e-08} & \textbf{3.561209e-06} \\ \cline{1-1}
    \textbf{HEMAS with two operators} &  &  &  & 2.817972e-01 \\ \hline
    \textbf{Griewank 500} & \multicolumn{1}{l|}{\textbf{EMAS}} & \multicolumn{1}{l|}{\textbf{HEMAS with one operator}} & \multicolumn{1}{l|}{\textbf{HEMAS with two operators}} & \textbf{HEMAS with three operators} \\ \hline
    \textbf{EMAS} &  & \textbf{9.068227e-04} & \textbf{3.865665e-16} & \textbf{7.564308e-16} \\ \cline{1-1}
    \textbf{HEMAS with one operator} &  &  & \textbf{1.401772e-06} & \textbf{2.104693e-06} \\ \cline{1-1}
    \textbf{HEMAS with two operators} &  &  &  & 9.349253e-01 \\ \hline
    \textbf{Griewank 1000} & \multicolumn{1}{l|}{\textbf{EMAS}} & \multicolumn{1}{l|}{\textbf{HEMAS with one operator}} & \multicolumn{1}{l|}{\textbf{HEMAS with two operators}} & \textbf{HEMAS with three operators} \\ \hline
    \textbf{EMAS} &  & \textbf{5.202443e-04} & \textbf{1.788370e-16} & \textbf{7.704896e-17} \\ \cline{1-1}
    \textbf{HEMAS with one operator} &  &  & \textbf{1.885065e-06} & \textbf{1.141232e-06} \\ \cline{1-1}
    \textbf{HEMAS with two operators} &  &  &  & 9.201805e-01 \\ \hline
    \textbf{Griewank 2000} & \multicolumn{1}{l|}{\textbf{EMAS}} & \multicolumn{1}{l|}{\textbf{HEMAS with one operator}} & \multicolumn{1}{l|}{\textbf{HEMAS with two operators}} & \textbf{HEMAS with three operators} \\ \hline
    \textbf{EMAS} &  & \textbf{8.371407e-04} & \textbf{1.579619e-16} & \textbf{2.894632e-17} \\ \cline{1-1}
    \textbf{HEMAS with one operator} &  &  & \textbf{9.102368e-07} & \textbf{3.212625e-07} \\ \cline{1-1}
    \textbf{HEMAS with two operators} &  &  &  & 8.411578e-01 \\ \hline
    \end{tabular}%
    }
	\label{tab:Dunn2}
\end{table}

\section{Conclusions}
In this paper, we have presented comprehensive results of the application of a novel hybridization concept in agent-based computing; namely, the autonomous hybridization of EMAS with selected classic metaheuristics. This approach assumes that a number of agents decide whether to be subjected to a selected hybrid optimization algorithm based on a predefined rule. Thus, the most important feature of agency (already visible in EMAS) becomes enhanced to a greater extent -- now, the agents can autonomously decide whether to undergo a selected hybridization step, thus enhancing their solutions by employing classic metaheuristics such as GA or PSO.

Three rules (based on measuring the diversity of the solution and the energy of the agents) were tested in the presented setting, and three versions of HEMAS produced results that were shown and discussed. The presented results pointed out that introducing two hybrid operators allowed for upgrading the efficiency and efficacy of EMAS; however, it seems that introducing a third operator did not bring any further enhancement in the tested cases.

In order to maintain the homogeneity of the whole algorithm, dedicated methods for redistributing the energy (a resource that controlled EMAS) were applied after returning from the hybrid step \cite{Redistribution_Godzik_2021}.

The discussed setting assumes the implementation of various deterministic rules (three selected ones were tested) for selected multi-dimensional problems. The observations proved that introducing the proposed hybridizations into EMAS along with the proper use of an energy-redistribution mechanism increased the efficiency and efficacy of the search with HEMAS as compared to EMAS. In particular, HEMAS with two hybridizations was better than the one with one (and without one) and not significantly worse than the one with three. Thus, it turned out that the two-hybrid version was sufficiently better than the one without hybridization for the setting that was presented in this paper.

Following the results presented in \cite{ICCCI2020}, EMAS and HEMAS additionally achieved better results faster than classic algorithms (e.g., the genetic algorithm). This allows us to claim that a hybridization of EMAS that is based on deterministic rules paves the way for the further development of a new promising metaheuristic algorithm that is built on EMAS (whose efficacy was proven theoretically) \cite{biul1}.

In this paper, we concentrated on common benchmarks that are used in metaheuristic-related research (multi-dimensional mathematical problems \cite{benchmarki}). The fact that HEMAS is built on EMAS (cf. \cite{biul1}) suggests that it may be applied for any similar problem; however, designers of such a system should be aware of the intricacies of each of its building blocks (as HEMAS is a complex algorithm that relies heavily on its hybrid parts). Here is an example: even though GA is an universal optimization technique and any possible problem may be approached by using this algorithm (providing that the proper representation is used and appropriate variation operators are selected), solving certain problems (e.g., discrete ones) may be tricky if one chooses PSO as the hybridization algorithm, as this algorithm is not inherently applicable to such problems (though certain modifications exist that can be used outright -- see, e.g., \cite{Clerc2004}). Moreover, the initialization of a hybrid step will not be straightforward when using an algorithm that relies on a completely different representation (e.g., ant colony optimization); however, one can again turn to detailed strategies that make such attempts possible \cite{DBLP:journals/ci/PolnikSBK21}.

In future work, we will also apply machine-learning methods as a decision factor for running certain hybrid optimization algorithms. Now, agents apply the mentioned rules; in the future, however, they will use a pre-trained machine-learning model (like a neural network, for example) which will serve as a decision factor for applying to start a hybrid optimization step.

\section*{Acknowledgments}
The research that is presented in this paper was partially supported by funds from the Polish Ministry of Education and Science that were assigned to AGH University of Science and Technology.
This research was supported in part by PL-Grid Infrastructure.

%\section*{References}

\bibliography{bibliography}

\end{document}